\documentclass{article}

\usepackage[main, final]{neurips_2026}

\usepackage[utf8]{inputenc} 
\usepackage[T1]{fontenc}    
\usepackage{hyperref}       
\usepackage{url}            
\usepackage{booktabs}       
\usepackage{amsfonts}       
\usepackage{nicefrac}       
\usepackage{microtype}      
\usepackage{xcolor}         

\usepackage{titletoc}
\usepackage{orcidlink}
\usepackage{graphicx}
\usepackage{amsmath}
\usepackage{amssymb}
\usepackage{amsthm}
\usepackage{dsfont}
\usepackage{algorithm}
\usepackage{algpseudocode}

\DeclareMathOperator*{\argmin}{arg\,min}
\newtheorem{theorem}{Theorem}
\newtheorem{lemma}{Lemma}
\newtheorem{proposition}{Proposition}
\newtheorem{observation}{Observation}

\newtheorem{definition}{Definition}
\newtheorem{problem}{Problem}
\newcommand{\spara}[1]{\smallskip\noindent{\bf #1}}

\title{Fair Bubble Sort: Provably Optimal Fair Ranking with Continuous Sensitive Attributes}

\author{%
 Antonio Ferrara~\orcidlink{0000-0002-0784-1115}\\
  Intesa Sanpaolo AI Research\\
  Turin, Italy\\
  \texttt{antonio.ferrara@intesasanpaolo.com} \\
  \And
 Fabio Vitale~\orcidlink{0009-0000-7398-9436}\\
  Intesa Sanpaolo Innovation Center\\
  Turin, Italy\\
  \texttt{fabio.vitale@intesasanpaolo.com} \\ 
    \And
  Atsushi Miyauchi~\orcidlink{0000-0002-6033-6433}\\
  Intesa Sanpaolo\\
  Turin, Italy\\
  \texttt{atsushi.miyauchi@intesasanpaolo.com} \\
    \And
  Francesco Bonchi~\orcidlink{0000-0001-9464-8315}\\
  Intesa Sanpaolo AI Research\\
  Turin, Italy\\
  \texttt{francesco.bonchi@intesasanpaolo.com} \\
}

\begin{document}

\maketitle

\begin{abstract}

Most fair-ranking methods assume categorical protected attributes, requiring continuous attributes such as age or income to be discretized into groups. This introduces arbitrary thresholds, loses fine-grained information, and ignores the natural ordering of the attribute. In this paper, we study the problem of fair ranking with continuous protected attributes without relying on thresholding. We introduce Continuous Pairwise Parity (CPP), a fairness notion based on Kendall correlation between ranking positions and the protected attribute, and seek rankings that satisfy a prescribed CPP constraint while minimizing their Kemeny distance from an initial score-based ranking. To solve this problem, we propose Fair Bubble Sort (FBS), an efficient post-processing algorithm based on targeted adjacent swaps. We prove that FBS is exactly optimal for the unweighted Kemeny distance and provide approximation guarantees for the score-weighted setting. FBS is output-sensitive, with running time determined by the number of swaps required to meet the fairness constraint. Experiments on synthetic and real-world datasets show that FBS is scalable and consistently outperforms discretization-based baselines, achieving a superior fairness-accuracy Pareto front.

\end{abstract}

\section{Introduction}

Algorithmic ranking systems increasingly influence decisions in high-stakes domains ranging from hiring to criminal justice, from credit scoring to college admission.
As a consequence, fairness aspects in algorithmic ranking have attracted growing attention~\cite{Jag19b,algfairnessdef2021,pitoura2022fairness,zehlike2022fairness,fairrankingcrticial2022}.
Much of such literature
deals with \emph{group fairness} along the lines of \emph{demographic parity}~\cite{dworkfairness2012} or \emph{equal opportunity}~\cite{HardtPNS16}, i.e.,
requiring the same chances of a positive outcome for different demographic groups, as defined by protected attributes, such as gender or race. In the case of ranking, this is
 typically cast as the problem of maximizing utility under a \emph{group-fairness constraint}, which requires a minimum fraction of individuals from each group to be included in the top-$k$ positions~\cite{Zehlike2017, CelisSV18,SinghJ18}.
The very notion of groups implies an underlying categorical attribute.
However, in many real-world scenarios, sensitive attributes are inherently continuous, such as age, income, or health scores. When dealing with such variables, the standard practice of discretizing them into groups introduces arbitrary thresholds, discards fine-grain information about the population (individuals falling into the same bucket are treated equivalently), and destroys the natural ordinality of the data (e.g., the fact that the bucket of age [20,30) comes after the bucket [13,20) and before the [30,45) one).
Despite recent progress on continuous sensitive attributes in classification and regression, fair ranking with continuous attributes remains largely unexplored. The core challenge lies in the combinatorial nature of rankings: standard continuous fairness regularizers cannot be trivially applied to listwise utility metrics or discrete ranking permutations. 

In this paper, we study the problem of producing a ranking that is as close as possible to a given score-based ranking while satisfying a fairness constraint defined directly on a continuous protected attribute, without any discretization. Our starting point is a pairwise view of fairness: in a fair ranking, an item with a larger protected-attribute value should be no more likely to rank above another item than below it. We capture this principle through \textbf{Continuous Pairwise Parity (CPP)}, defined using the Kendall correlation between ranking positions and the protected attribute. CPP requires no discretization, preserves the ordinal information in the attribute, and reduces to pairwise statistical parity for binary protected attributes.

Given a score-based ranking, our objective is to find the closest ranking satisfying a prescribed CPP constraint. We measure distortion using the Kemeny distance, as it mirrors the pairwise structure of the fairness notion,
decomposing the utility loss as a sum over pairs of items whose order in the produced ranking disagrees with the score-based ranking. Weighting each pair by their difference in score further captures the intuition that swapping items of near-identical merit is essentially costless, while swapping items of very different scores is expensive. This shared pairwise decomposition of cost and fairness is precisely what allows a single algorithmic primitive -- the adjacent swap -- to navigate both quantities, and is the structural feature on which our theoretical guarantees rest (see Appendix~\ref{app:motivations} for an extended discussion motivating Continuous Pairwise Parity).

\spara{Paper Contributions.} The main contributions of this paper are summarised as follows:

\begin{enumerate}
    \item We introduce the fairness notion of Continuous Pairwise Parity (CPP), extending \emph{pairwise statistical parity} \cite{ narasimhan2020pairwise} to continuous protected attributes. 
    \item We propose \textbf{Fair Bubble Sort (FBS)}, the first theoretically grounded post-processing algorithm that optimally navigates the utility-fairness trade-off for continuous attributes via targeted adjacent swaps.
    \item We prove that FBS is \emph{exactly optimal} for the unweighted Kemeny distance (Theorem~\ref{thm:unweighted-optimality}). For the weighted case, we prove an asymptotic ratio of $(1{+}\sqrt{2})/2\approx 1.21$ for integer scores in the all-ascending attribute case with $n$ a power of two (Proposition~\ref{prop:ascending-bound}). Extensive empirical evidence further suggests that FBS may satisfy a general $2$-approximation bound.
    \item We show that FBS is highly scalable, running in $O(n\log n+K^{\ast})$ time for the unweighted case and in $O\!\bigl((n+K^{\ast})\log n\bigr)$ time for the weighted case, where $K^{\ast}$ is the minimum number of adjacent swaps required to satisfy the fairness constraint (Propositions~\ref{prop:complexity}).
    \item We validate FBS on four real-world datasets, empirically demonstrating its scalability and its superior performance over non-trivial baseline models.
\end{enumerate}

\section{Related Work}
\label{sec:related_work}

Our work sits at the intersection of algorithmic fair ranking and the treatment of continuous sensitive attributes, bridging two traditionally separate lines of research.

\spara{Fair Ranking.}
The literature on fair ranking has grown significantly, driven by the need to mitigate historical biases in search, recommendation, selection, and allocation systems \cite{Jag19b,algfairnessdef2021,pitoura2022fairness,zehlike2022fairness,fairrankingcrticial2022}.
With a few exceptions that focus on individual fairness~\cite{dworkfairness2012,biega2018equity,saito2022fair}, or that tackle individual and group fairness jointly \cite{2021maxminkdd, zehlike2020matching},  vast majority of the approaches 
formulate fairness as a group-level constraint, assuming that the protected attribute is categorical (e.g., binary gender or discrete race). Solutions typically enforce variants of demographic parity or proportional representation, ensuring that a minimum fraction of items from each group appears in the top-$k$ positions or in every prefix of the ranking \cite{YangS17,CelisSV18,Zehlike2017,SinghJ18,YangGS19,AsudehJS019,geyik2019fairness}.
Conceptually closest to our work is the notion of \emph{pairwise fairness} \cite{kuhlman2020rank, narasimhan2020pairwise}, which evaluates the probability of a protected candidate being ranked above a non-protected one. A fundamental limitation of all these approaches is their reliance on predefined protected groups, requiring either categorical protected attributes or discrete bins. When dealing with inherently continuous protected attributes, enforcing these metrics requires arbitrary thresholding, which discards fine-grained intra-group variance and destroys the ordinal nature of the data.

\spara{Continuous Sensitive Attributes.}
In the broader machine learning research, the flaws of discretizing continuous sensitive attributes have been recently recognized, primarily within the contexts of regression and classification. Seminal works have proposed matching continuous distributions or penalizing statistical dependence to enforce fairness. For instance, \cite{pmlr-v97-mary19a} introduced fairness-aware learning using the Rényi maximum correlation coefficient, while \cite{pmlr-v80-komiyama18a} formulated non-convex optimization techniques for fair regression. Another prominent line of research leverages causal inference, where continuous sensitive attributes are naturally treated as nodes in structural equation models. In this context, \cite{kusner2017counterfactual} introduced \emph{counterfactual fairness}, ensuring that predicted distributions remain invariant under counterfactual changes to the continuous protected attribute, a framework later extended by \cite{Chiappa19} to isolate path-specific fair effects.
Alternatively, frameworks based on \emph{individual fairness} \cite{dworkfairness2012} accommodate continuous spaces by requiring that similar individuals receive similar predictions.
While mathematically robust, these continuous regularizers, segmentation strategies, and causal frameworks are designed for differentiable loss functions and point-wise predictions, making them inapplicable to the discrete, combinatorial nature of listwise ranking systems.

\spara{Bridging the Gap.}
The specific intersection of fair ranking and continuous protected attributes is still largely unexplored. While recent preliminary work by Giuliani et al. \cite{GiulianiMC024} has highlighted the necessity of moving beyond discrete groups by exploring continuous attributes within long-term, dynamical allocation systems, the fundamental algorithmic challenge of single-shot, static fair ranking remains unresolved. 
To the best of our knowledge, we propose the first theoretically grounded post-processing algorithm to enforce fairness over continuous attributes in static ranking.

\section{Preliminaries}\label{sec:preliminaries}

Let $U=[n]$ be a set of $n$ items and $S_U$ be the symmetric group of all permutations of~$U$. A ranking $r$ is an element of $S_U$, i.e., a bijection $r:U\to[n]$, where $r(u)$ denotes the position of item $u$ in the ranking. Each item $u\in U$ has a continuous \textbf{sensitive attribute} $a(u)\in\mathbb{R}_+$ and a \textbf{score} $s(u)\in\mathbb{R}_+$. The score $s(u)$ indicates the overall fitness/utility of $u$ for the task at hand: it might be, e.g., the results of a test in a college admission, or the output of a complex machine learning model in a hiring setting. We assume higher score $s$ is better. Ranking simply by $s$ is not considered adequate in the fairness literature, as $s$ might encode historical biases, which fair ranking aims to fix by producing a ranking that is as close as possible to the one induced by $s$ while meeting the given fairness criteria. We denote $r_s$ the ``unfair'' \textbf{score-based ranking} obtained by sorting items in descending order of~$s$.

\spara{Kemeny distance.} We denote by
$
\mathrm{Inv}(r_s,r):=\{(i,j) \mid 1\le i<j\le n,\,(r_s(j)-r_s(i))(r(i)-r(j))>0\}
$
the inversion set of $r$ w.r.t. $r_s$, the item pairs whose relative order in $r$ differs from that in $r_s$. We denote by $d(r_s,r)$ the Kemeny distance\footnote{The number of pairs that two rankings order oppositely is also known as the Kendall-$\tau$ distance~\cite{fagin2004comparing}; Kemeny and Snell~\cite{kemeny1962preference} derived this distance from a set of axioms, and the related term ``Kemeny ranking'' denotes the ranking that minimizes the total distance to a given set of rankings~\cite{conitzer2006improved}. Our $d(r_s,r)$ is a normalized version of this distance in which each inverted pair carries a weight. We call it the Kemeny distance to keep it distinct from the Kendall-$\tau$ \emph{correlation} that defines the fairness measure below.} between $r_s$ and an alternative ranking~$r$, defined as
\[
d(r_s,r)=\sum_{(i,j)\in\mathrm{Inv}(r_s,r)} w_{ij},
\]
where $w_{ij}\ge 0$ is a non-negative weight assigned to each pair. While the proposed algorithm and all theoretical results hold for any weight function $w_{ij}\ge 0$, in this paper we consider two cases: the \emph{unweighted} case with uniform weights $w_{ij}=1/\binom{n}{2}$ and the \emph{score-gap weighted} case  with $w_{ij}=|s(i)-s(j)|/W$, where $W =\sum_{1\le k<l\le n}|s(k)-s(l)|$ is a normalization constant. Both choices yield $d\in[0,1]$, but in the score-gap weighted case swapping items with closer scores is cheaper than swapping items with further scores.

\spara{Continuous Pairwise Parity (CPP).} Let $N_{\nearrow}(r) := \sum_{1\leq i<j\leq n} \mathds{1}\bigl[(r(j) - r(i))(a(j) - a(i)) > 0\bigr]$ and $N_{\searrow}(r) := \sum_{1\leq i<j\leq n} \mathds{1}\bigl[(r(j) - r(i))(a(j) - a(i)) < 0\bigr]$ denote the number of \emph{concordant} and \emph{discordant} pairs between the ranking and the attribute values, respectively. The \emph{Kendall}-$\tau$ \emph{correlation} between a ranking $r$ and the attribute~$a$ is
$
\tau(r,a) = \frac{N_{\nearrow}(r) - N_{\searrow}(r)}{\binom{n}{2}}.
$
A ranking is fair when this correlation is close to zero. Since $\tau(r,a)\in[-1,1]$, we define the CPP fairness measure as
$
\phi(r):=1-|\tau(r,a)|,
$
so that $\phi(r)=1$ indicates perfect fairness and $\phi(r)=0$ maximal unfairness. CPP needs no binning of the attribute, uses only its order, and decomposes over pairs in the same way as the Kemeny distance; when the attribute is binary, $\phi(r)=1$ is exactly the pairwise statistical parity of~\cite{kuhlman2020rank} (Appendix~\ref{app:motivations}).


\begin{problem}\label{prob:main}
Given the set of $n$ items $U$, their sensitive attribute $a$ and score $s$, given the weighted or unweighted Kemeny distance $d(\cdot,\cdot)$ and the CPP fairness measure $\phi(\cdot)$, as defined above, find the ranking~$r$ that minimizes the distance from the score-based ranking $r_s$, subject to the fairness constraint $\phi(r) \ge \psi$, where $\psi\in[0,1]$ is a predefined fairness threshold:
\begin{equation}
    \begin{aligned}
        \argmin_{r \in S_U \, : \, \phi(r) \ge \psi} \quad & d(r_s, r).
    \end{aligned}
    \label{eq:optimization_problem}
\end{equation}
\end{problem}

\spara{Remark.}
Throughout the paper we assume that $s$ and $a$ are both injective, i.e.\ no two items share the same score or the same attribute value. This is without loss of generality: whenever $s(u)=s(v)$, the pair $\{u,v\}$ contributes $|s(u)-s(v)|=0$ to the weighted Kemeny distance, so swapping them is free; whenever $a(u)=a(v)$, the pair contributes to neither $N_{\nearrow}$ nor $N_{\searrow}$, so their ordering does not affect~$\phi$. Ties can therefore be broken by an arbitrary total order, without losing the theoretical results presented in the paper (see detailed discussion at the end of Section 4).

\section{Method}
Both our notions of fairness and distance from the score-based ranking are defined in terms of pairwise comparisons. This shared structure suggests a natural algorithmic primitive: the \emph{adjacent swap}. Swapping two items at consecutive positions $k,k{+}1$ flips the orientation of exactly one pair -- it either creates or resolves a single inversion with respect to~$r_s$, and it changes the difference $N_{\nearrow}-N_{\searrow}$ by exactly~$\pm 2$ (Observation~\ref{obs:single-swap}). No other pair is affected. This locality makes adjacent swaps the atomic unit of both cost and fairness progress: each swap incurs the smallest possible perturbation to the ranking while making a discrete, predictable step toward the fairness target. In the following, for convenience, we refer to $\mathrm{sign}(N_{\nearrow}-N_{\searrow})$
as the \emph{majority sign} and we work with the \emph{imbalance} $\Delta(r):=|N_{\nearrow}(r)-N_{\searrow}(r)|$, so that the constraint $\phi(r)\ge\psi$ is equivalent to
\[
\Delta(r)\le\Delta^{\ast}:=\Bigl\lfloor\tfrac{n(n-1)}{2}(1-\psi)\Bigr\rfloor.
\]
The key design insight is that, when $\Delta(r)>\Delta^{\ast}$, there always exists at least one adjacent pair of the majority sign whose swap reduces $\Delta$ by~$2$ (Lemma~\ref{lem:valid-swap-exists}). A greedy strategy that repeatedly picks such a swap is therefore always well defined; moreover, since each swap reduces $\Delta$ by exactly~$2$, the total number of swaps $K^{\ast}$ is determined by the initial imbalance $\Delta_0:=\Delta(r_s)$ and the threshold alone. Note that $\Delta$ retains the parity of~$\Delta_0$; the algorithm therefore sets $\Delta^{\ast}\gets\max\!\bigl(\Delta^{\ast},\,\Delta_0\!\bmod 2\bigr)$ to guarantee feasibility. Every swap made by FBS improves fairness and increases the distance from $r_s$ (see the proof of Theorem~\ref{thm:unweighted-optimality}), so FBS never moves back towards $r_s$. One may ask whether this costs optimality, that is, whether some optimal ranking can only be reached by paths that at some point move back towards $r_s$. Lemma~\ref{lem:nondecreasing} in Section~\ref{sec:proof-optimality} shows that it does not: every optimal ranking is reachable from $r_s$ by adjacent swaps along which the distance from $r_s$ never decreases.

\begin{algorithm}[h!]
\caption{Fair Bubble Sort (FBS) -- unweighted Kemeny distance}\label{alg:fbs-unweighted}
\begin{algorithmic}[1]
\Require Items $U = [n]$, attribute $a: U \to \mathbb{R}_+$, score $s: U \to \mathbb{R}_+$, fairness threshold $\psi\in[0,1]$
\Ensure Ranking $R$ solving Problem \ref{prob:main} with the unweighted Kemeny distance
\State $(R,A)\gets$ arrays initialized according to $r_s$ (descending order of score) \Comment{initial $r_s$}
\State $\sigma[k]\gets\mathrm{sign}(A[k+1]-A[k])$ for $k=1,\ldots,n-1$ \Comment{adjacent-pair signs}
\State $N_{\nearrow}\gets |\{i<j:A[i]<A[j]\}|$;\quad $N_{\searrow}\gets |\{i<j:A[i]>A[j]\}|$
\State $\Delta^{\ast}\gets\max\!\bigl(\lfloor(1-\psi)\,n(n-1)/2\rfloor,\;|N_{\nearrow}-N_{\searrow}|\!\bmod 2\bigr)$ \Comment{parity-safe threshold}
\State $\mu\gets\mathrm{sign}(N_{\nearrow}-N_{\searrow})$ \Comment{majority sign}
\State $L\gets$ doubly-linked list of $\{k:\sigma[k]=\mu\}$ \Comment{build in $O(n)$}
\While{$|N_{\nearrow}-N_{\searrow}|>\Delta^{\ast}$}
    \State $k^{\ast}\gets$ any element of $L$ \Comment{$O(1)$}
    \State swap $R[k^{\ast}]\leftrightarrow R[k^{\ast}+1]$, $A[k^{\ast}]\leftrightarrow A[k^{\ast}+1]$
    \For{$\ell\in\{k^{\ast}-1,\;k^{\ast},\;k^{\ast}+1\}\cap\{1,\ldots,n-1\}$}
        \State $\sigma_{\mathrm{old}}\gets\sigma[\ell]$
        \If{$\ell=k^{\ast}$}
            \State $\sigma[\ell]\gets-\sigma[\ell]$ \Comment{middle sign flips}
        \Else
            \State $\sigma[\ell]\gets\mathrm{sign}(A[\ell+1]-A[\ell])$
        \EndIf
        \If{$\sigma_{\mathrm{old}}=\mu$ \textbf{and} $\sigma[\ell]\ne\mu$}
            \State remove $\ell$ from $L$ \Comment{$O(1)$}
        \ElsIf{$\sigma_{\mathrm{old}}\ne\mu$ \textbf{and} $\sigma[\ell]=\mu$}
            \State insert $\ell$ into $L$ \Comment{$O(1)$}
        \EndIf
    \EndFor
    \State $N_{\nearrow}\gets N_{\nearrow}-\mu$;\quad $N_{\searrow}\gets N_{\searrow}+\mu$
\EndWhile
\State \Return $R$
\end{algorithmic}
\end{algorithm}

The resulting algorithm, dubbed \textbf{Fair Bubble Sort (FBS)}, is provided in Algorithm~\ref{alg:fbs-unweighted} for the unweighted Kemeny distance. The pseudocode for the weighted case is instead given in Algorithm~\ref{alg:fbs-weighted} in Appendix~\ref{app:weighted-alg}. Like classical bubble sort, our method operates exclusively via adjacent swaps, selecting at each step a swap that improves fairness.
In the pseudocode we use uppercase arrays indexed by position: $R[i]$ is the item at position~$i$ and $A[i]$ its attribute (i.e., $R$ stores the ranking in one-line notation, the inverse of the function $r:U\to[n]$ defined in Section~\ref{sec:preliminaries}).
Lines~1--6 initialise the algorithm. The arrays are initialized according to the score-based ranking~$r_s$ (descending order of score, line~1). The sign array~$\sigma$ records the orientation of each adjacent pair (line~2), and the global counts $N_{\nearrow}$, $N_{\searrow}$ are computed in $O(n\log n)$ via a merge-sort--based count (line~3).
The parity-safe threshold $\Delta^{\ast}$ is derived from~$\psi$ (line~4), the majority sign~$\mu$ identifies which type of pair is in excess (line~5), and the linked list~$L$ collects all positions eligible for swapping (line~6).

The main loop (lines~7--24) iterates until the imbalance $\Delta(r)$ falls within the threshold~$\Delta^{\ast}$.
At each iteration, an arbitrary eligible position~$k^{\ast}$ is retrieved from~$L$ in $O(1)$ (line~8) and the corresponding adjacent pair is swapped (line~9).
The for-loop (lines~10--22) performs the local update: it examines the three sign entries that may have changed -- $\sigma[k^{\ast}{-}1]$, $\sigma[k^{\ast}]$, $\sigma[k^{\ast}{+}1]$ -- and updates $L$ accordingly.
The global counts are updated in $O(1)$ (line~23).

\paragraph{Unweighted vs.\ weighted Kemeny distance.}
Algorithm~\ref{alg:fbs-unweighted} presents FBS for the unweighted Kemeny distance, where every swap has unit cost. Since the total cost depends only on the \emph{number} of swaps $K^{\ast}$ and not on which majority-sign pairs are selected (Theorem~\ref{thm:unweighted-optimality}), any majority-sign adjacent pair is equally good, and a doubly-linked list~$L$ suffices to maintain the pool of eligible positions in $O(1)$ per swap.
When the Kemeny distance is weighted by score gaps $w_{ij}=|s(i)-s(j)|$, different swaps have different costs and FBS must select the cheapest one at each step. The algorithm additionally maintains a score array $S[i]$; the linked list is replaced by a min-heap keyed by the adjacent cost $c[k]=|S[k]-S[k+1]|$, yielding an $O((n+K^{\ast})\log n)$ algorithm (the full pseudocode is given in Algorithm~\ref{alg:fbs-weighted} in Appendix~\ref{app:weighted-alg}).

\spara{Handling ties in the attribute.}
Both algorithms (Algorithms~\ref{alg:fbs-unweighted} and~\ref{alg:fbs-weighted}) are presented, for simplicity sake, under the general-position assumption that attribute values are pairwise distinct. In practice, real-valued attributes may contain ties (e.g.\ integer-valued age or discretised scores). Two modifications suffice to handle ties correctly; the algorithm structure, the greedy optimality proof, and the complexity bounds are otherwise unchanged.

\begin{enumerate}
\item \textbf{Adjacent signs.}
The sign array becomes $\sigma[k]\in\{+1,0,-1\}$, where $\sigma[k]=0$ when $A[k]=A[k+1]$.
Swapping a tied pair does not change the orientation of that pair---it was tied before and remains tied after---so $\Delta(r)$ is unaffected.
Positions with $\sigma[k]=0$ are therefore never inserted into $L$ (or~$H$) and are never selected for swapping.

\item \textbf{Neighbour updates.}
After a swap at position~$k^{\ast}$, the recomputed sign $\sigma[\ell]$ for a neighbour $\ell\in\{k^{\ast}{-}1,\,k^{\ast}{+}1\}$ may transition to or from~$0$.
If $\sigma[\ell]$ was~$\mu$ and becomes~$0$, the entry is removed from $L$ (or~$H$); if it was~$0$ and becomes~$\mu$, it is inserted.
The middle entry $\sigma[k^{\ast}]$ flips sign as before ($+1\leftrightarrow -1$) and cannot become~$0$, since the swap reverses a strict inequality.
\end{enumerate}

All remaining logic -- the imbalance $\Delta(r)=|N_{\nearrow}-N_{\searrow}|$, the threshold~$\Delta^{\ast}$, the majority sign~$\mu$, and the $O(1)$ update of $N_{\nearrow},N_{\searrow}$ per swap -- carries over without change.


\section{Optimality of FBS for the Unweighted Problem}\label{sec:proof-optimality}
The proof of optimality builds on four results: $(i)$~an adjacent swap changes the imbalance by exactly $\pm 2$ and the number of inversions by exactly $\pm 1$ (Observation~\ref{obs:single-swap}); $(ii)$~any ranking can be realized by adjacent swaps only, with the shortest sequence of swaps having length equal to $|\mathrm{Inv}(r_s,r)|$ (Lemma~\ref{lem:contiguous}); $(iii)$~a valid majority-sign swap always exists when the imbalance exceeds the threshold (Lemma~\ref{lem:valid-swap-exists}); $(iv)$~FBS uses the minimum number of swaps among all feasible solutions (Proposition~\ref{prop:minimal-swaps}). A further structural result motivates the design of FBS: every optimal ranking is reachable from $r_s$ by adjacent swaps along which the distance from $r_s$ never decreases (Lemma~\ref{lem:nondecreasing}), so the search space of algorithms that never move back towards $r_s$, to which FBS belongs, still contains every optimal ranking.

%
Throughout this subsection, an adjacent swap action $\tau_{k}$ exchanges positions $k$ and $k{+}1$ in a ranking $r$; an \emph{algorithm} $\pi$ is an ordered sequence of adjacent swaps $(\tau_{t_1}, \ldots, \tau_{t_m})$ such that their composition $\pi = \tau_{t_m}\circ\cdots\circ\tau_{t_1}$ determines the output ranking $\pi\circ r_s$. 
The intermediate rankings produced by the sequence of swaps are denoted $r^{(0)}=r_s,\; 
r^{(1)}= \tau_{t_1} \circ r^{(0)},\; \ldots, r^{(m)}=\tau_{t_m}\circ r^{(m-1)}$. We let $A_{swap}$ denote the set of all such algorithms and $A_{swap,\nearrow}\subseteq A_{swap}$ the subset of algorithms such that $d(r_s,r^{(t)})$ is non-decreasing in~$t$. A key elementary fact used repeatedly below is the following.

%


\begin{observation}[Single-swap increment]\label{obs:single-swap}
Let $r'$ be obtained from $r$ by an adjacent swap of the items $u,v$ at positions $k,k+1$, $r' = \tau_k \circ r$. Then
$
d(r_s,r')-d(r_s,r)=\pm\,w_{uv},
$
where the sign is $+$ iff the swap creates an inversion with respect to $r_s$ and $-$ iff it resolves one. Moreover, the same swap flips the Kendall orientation of the pair $\{u,v\}$ between $\nearrow$ and $\searrow$ and leaves all other pair orientations unchanged, so
\[
\Delta(r')-\Delta(r)\in\{-2,+2\}.
\]
\end{observation}

\begin{lemma}[Reduction to adjacent swaps~\cite{knuth1998art}]\label{lem:contiguous}
For every ranking $r\in S_U$ there exists an algorithm $\pi\in A_{swap}$ such that $\pi\circ r_s=r$, and the shortest such algorithm has length $|\mathrm{Inv}(r_s,r)|$.
\end{lemma}

\begin{lemma}[Reduction to non-decreasing swaps]\label{lem:nondecreasing}
Let $r^{\ast}$ be an optimal solution to Problem \ref{prob:main}, then there exists an algorithm $\pi\in A_{swap,\nearrow}$ such that $\pi\circ r_s=r^{\ast}$.
\end{lemma}

\begin{lemma}[Existence of a valid greedy swap]\label{lem:valid-swap-exists}
For any $r \in S_U$, consider $b_k:=a(r^{-1}(k))$ for $k=1,\ldots,n$, where $r^{-1}(k)\in U$ is the item at rank $k$. Under the injectivity assumption of $a(\cdot)$, the sequence $b_1,\ldots,b_n$ consists of pairwise distinct reals. Hence, a pair of ranks $(i,j)$ with $i<j$ is $\nearrow$ iff $b_i<b_j$ and $\searrow$ iff $b_i>b_j$. If $N_{\nearrow}(r)\ne N_{\searrow}(r)$, then $r$ contains at least one adjacent pair of the majority sign, i.e.\ an index $k\in\{1,\ldots,n-1\}$ whose flip strictly reduces $\Delta(r)$ by $2$.
\end{lemma}

\begin{proposition}[Minimality of the number of greedy swaps]\label{prop:minimal-swaps}
Assume without loss of generality $N_{\nearrow}(r_s)\ge N_{\searrow}(r_s)$ and set $\Delta_0:=\Delta(r_s)$. If $\Delta_0>\Delta^{\ast}$, then FBS is well defined and terminates after exactly $K^{\ast}:=\lceil(\Delta_0-\Delta^{\ast})/2\rceil$ swaps. Moreover, any algorithm in $A_{swap}$ that outputs a feasible ranking uses at least $K^{\ast}$ swaps.
\end{proposition}


\begin{theorem}[Optimality of FBS for the unweighted Kemeny distance]\label{thm:unweighted-optimality}
Let $r_{\mathrm{FBS}}$ denote the ranking returned by FBS. Then $r_{\mathrm{FBS}}$ is an optimal solution of Problem~\ref{prob:main} for the unweighted Kemeny distance. Moreover, $d(r_{\mathrm{FBS}},r_s) = K^{\ast}/\binom{n}{2}$ where $K^{\ast}=\lceil(\Delta_0-\Delta^{\ast})/2\rceil$ as in Proposition~\ref{prop:minimal-swaps}.
\end{theorem}

All proofs for the results above are given in Appendix~\ref{app:proofs}.


\section{Approximation Bound for the Weighted Problem}\label{sec:weighted-approx}
When the Kemeny distance is weighted, different swaps have different costs and FBS must select the cheapest majority-sign adjacent pair at each step.
Algorithm~\ref{alg:fbs-weighted} (see pseudocode in Appendix~\ref{app:weighted-alg}) replaces the doubly-linked list of Algorithm~\ref{alg:fbs-unweighted} with a min-heap keyed by the adjacent cost $c[k]=|S[k]-S[k+1]|$. Because FBS does not necessarily select the $K^{\ast}$ feasible pairs of smallest cumulative weight, it loses the optimality property of the unweighted case. Nonetheless, for any strictly positive weight function, we can bound its sub-optimality in terms of the weight spread. 

All the proofs of this section are given in Appendix~\ref{app:proofs_weighted}.

\begin{proposition}[Weight-spread approximation bound for FBS]\label{prop:weighted-approx}
Let $w_{ij}>0$ be an arbitrary strictly positive weight function, and let $r_{\mathrm{FBS}}$ be the ranking produced by FBS and $r^{\ast}$ an optimal solution to Problem~\ref{prob:main} for the corresponding weighted Kemeny distance. Let $w_{\min}:=\min_{i<j}w_{ij}$ and $w_{\max}:=\max_{i<j}w_{ij}$.
Then:
$d(r_s,r_{\mathrm{FBS}})\ \le\ w_{\max} / w_{\min} \, d(r_s,r^{\ast}).
$
\end{proposition}


In the remainder of this section, we study a natural and practically relevant setting in which the scores $s$ of the $n$ items are mapped to $[n]$ -- only the ordinal information of the score $s$ is maintained --
and the items are identified according to their score: the highest-score item will be identified with $n$, the second highest by $n-1$, and so on till the smallest-score item $1$. Having $s(i)=i$ for $i\in[n]$, score-gap weights reduce to positional gaps $w_{ij}=|i-j|/W$, penalizing large displacements more than local rearrangements. In this setting, Proposition~\ref{prop:weighted-approx} yields $w_{\max}/w_{\min}=n-1$, too loose to be informative. To obtain tighter bounds, we exploit the following identity, whose proof builds on a left/right inversion decomposition in the spirit of the displacement-inversion analysis of Diaconis and Graham~\cite{diaconis1977spearman}.

\begin{lemma}[Spearman--Kendall identity]\label{lem:spearman-kendall}
For integer scores $s(i)=i$, $i=1,\ldots,n$, with unnormalised pair weights $w_{ij}=|i-j|$ and any ranking~$r$,
\[
d(r_s,r)\ =\ \tfrac{1}{2W}\,S_2(r),\qquad\text{where}\quad S_2(r):=\sum_{i=1}^{n}\bigl(r(i)-i\bigr)^2 \text{is the Spearman squared displacement.}
\]
\end{lemma}

This identity transforms the weighted distance from a sum over $O(n^2)$ \emph{pairs} into a sum over $n$ \emph{items}, where each term depends only on how far item~$i$ has moved from its original position. The approximation ratio becomes a ratio of squared $\ell_2$ norms: $d(r_s,r_{\mathrm{FBS}})/d(r_s,r^{\ast})=\|\boldsymbol{\delta}_{\mathrm{FBS}}\|_2^2/\|\boldsymbol{\delta}^{\ast}\|_2^2$, where $\boldsymbol{\delta}_i=r(i)-i$. Exploiting this identity, we can derive the exact FBS cost in closed form for the worst-case attribute configuration (i.e.,  the \emph{all-ascending} case $a=s$
where all pairs are concordant and thus the initial imbalance is maximal), maximal fairness requirement
$\psi=1$, and $n$ a power of two.

\begin{proposition}[Tight bound for the all-ascending case]\label{prop:ascending-bound}
For integer scores $s(i)=i$ with $a=s$, $\psi=1$, and $n=2^t\ge 4$, FBS produces a ranking 
$r_{\mathrm{FBS}}$ such that:
\[
\frac{d(r_s,r_{\mathrm{FBS}})}{\mathrm{LP}}\ \longrightarrow\ \frac{1+\sqrt{2}}{2}\ \approx\ 1.207\qquad\text{as }n\to\infty,
\]
where $\mathrm{LP}$ is the LP relaxation lower bound, $\mathrm{LP} = \sum_{k=1}^{K^{\ast}}w_{(k)}$, where $w_{(1)}\le w_{(2)}\le\cdots\le w_{(K^{\ast})}$ are the sorted pair weights. In particular, $d(r_s,r_{\mathrm{FBS}})<2\,d(r_s,r^{\ast})$ for all~$n=2^t$ with $t\ge 2$.
\end{proposition}


The above asymptotic ratio $d_{\mathrm{FBS}}/\mathrm{LP}\to \frac{1+\sqrt{2}}{2}$ is \emph{independent of~$n$} and has been further verified computationally for all $n=2^t$ with $t=2,\ldots,21$ (Table~\ref{tab:ratio-appendix} in Appendix~\ref{app:lp-relaxation}).

Our analysis and experiments suggest that the $2$-approximation holds in general, dropping the \emph{all-ascending} assumption $a=s$. That is, we conjecture that $d(r_s,r_{\mathrm{FBS}})<2\,d(r_s,r^{\ast})$ for integer scores $s(i)=i$, for any attribute array~$a$, and any feasible~$\Delta^{\ast}$. Indeed, combining the Spearman--Kendall identity (Lemma~\ref{lem:spearman-kendall}) with the classical Diaconis--Graham inequality $K\le F\le 2K$~\cite{diaconis1977spearman} yields a natural proof target of factor~$2$, since the right-hand inequality $F\le 2K$ is tight in general. Tightening below~$2$ would require arguments beyond classical norm comparisons that exploit the specific structure of FBS's output (see Appendix~\ref{app:proof-strategy} for further discussion). We next report empirical evidence supporting the 2-approximation.

\begin{figure}[h!]
    \centering
    \includegraphics[width=.9\textwidth]{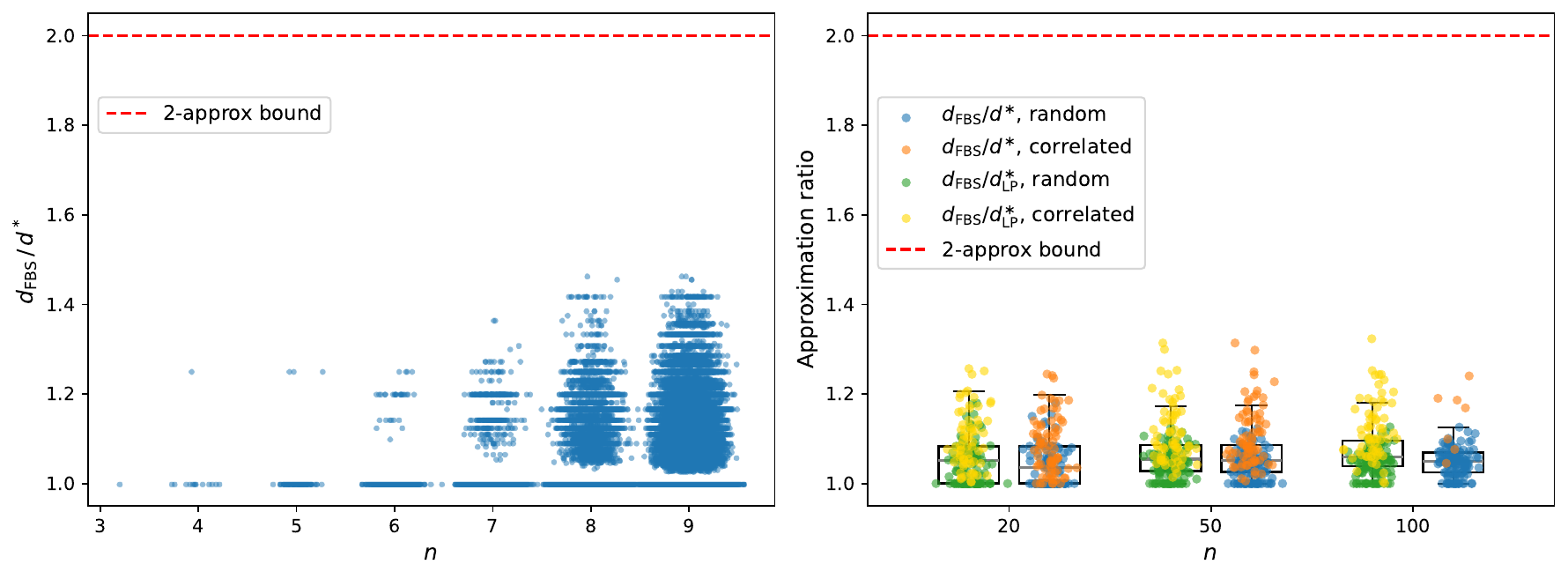}
    \vspace{-6mm}
    \caption{Empirical approximation ratio of FBS for the weighted Kemeny distance with integer scores $s(i)=i$ and fairness threshold $\Delta^{\ast}$ set to the tightest feasible value. \textbf{(a)}~Exact ratio $d_{\mathrm{FBS}}/d^{\ast}$ for all attribute permutations and all feasible $\Delta^{\ast}$, $n=3,\ldots,9$ ($699{,}282$ instances). \textbf{(b)}~For $n\in\{20,50,100\}$, MILP-exact ratio $d_{\mathrm{FBS}}/d^{\ast}$ (blue: random instances, orange: correlated instances) and LP-bound upper bound $d_{\mathrm{FBS}}/d^{\ast}_{\mathrm{LP}}$ (green: random, yellow: correlated); at each $n$, the left box summarizes the LP-bound ratios and the right box the MILP-exact ratios. The dashed red line marks the conjectured $2$-approximation bound.}
    \label{fig:approx-ratio}
        \vspace{-2mm}
\end{figure}


\spara{Brute-force exact computation for small instances.}
For $n=3,\ldots,9$ with integer scores $s(i)=i$, we enumerate every attribute permutation~$a$ and every feasible fairness threshold~$\Delta^{\ast}$ (all values with the correct parity). For each instance we compute the FBS cost~$d_{\mathrm{FBS}}$ and the exact optimum~$d^{\ast}$ by brute-force search over all $n!$ permutations, recording the ratio $d_{\mathrm{FBS}}/d^{\ast}$.
Figure~\ref{fig:approx-ratio}(a) shows a scatter plot of all 699{,}282 non-trivial instances across these values of~$n$. Every ratio lies strictly below~$2$, with the worst observed value $\approx 1.46$ at $n=9$, supporting the 2-approximation conjecture.

\spara{Exact MILP optimum for medium instances.}
For $n\in\{20,50,100\}$ we generate random and correlated attribute arrays and solve the integer programme exactly via MILP (binary inversion variables with FAS-style triangle constraints, see Appendix~\ref{app:milp}). We use HiGHS via \texttt{scipy.optimize.milp} with a $180$\,s solver time-limit per instance, a $0.1\%$ relative gap tolerance, and a $300$\,s wall-clock subprocess cap to handle pathologically hard correlated configurations. We also report the LP-relaxation lower bound $d^{\ast}_{\mathrm{LP}}\le d^{\ast}$ on $612$ instances, so that $d_{\mathrm{FBS}}/d^{\ast}_{\mathrm{LP}}$ provides a complementary upper-bound on the true approximation ratio.
Figure~\ref{fig:approx-ratio}(b) compares the two: the LP-bound and MILP-exact distributions coincide for the hardest correlated instances at $n=20$ and $n=50$ (the LP relaxation is tight there), but the MILP-exact ratio is strictly smaller for the random configurations at $n=100$, where the LP bound has slack. Across all $542$ MILP-solved instances the worst observed exact ratio is $1.31$, well below the conjectured $2$.
The ratio $d_{\mathrm{FBS}}/\mathrm{LP}$ for the all-ascending case $a=s$ (up to $n\approx 2\times 10^6$, including non-power-of-two values) is reported in Appendix~\ref{app:extra-experiments}.


\section{Comparison with Baselines}\label{sec:exp-baselines}
We next compare FBS against baselines on the quality of the ranking produced: as FBS is optimal in the unweighted case, we focus these experiments on the weighted case. Given that existing fair-ranking algorithms assume discrete protected groups, we discretize the continuous attribute into $G$ equal-frequency bins and apply two representative methods using their original implementations: \textbf{DetConstSort}~\citep{geyik2019fairness} (FairRankTune library~\citep{cachel2024fairranktune}, tested at $G\in\{2,3,5,10\}$) and \textbf{FA*IR}~\citep{Zehlike2017} (fairsearchcore library \cite{fairsearchcore}, with $G=2$ since FA*IR is inherently binary).
We also include two baselines of our own: a simple \textbf{median-split interleaving} that splits items at the median attribute value into two groups, sorts each by descending score, and alternately picks the next highest-scored item from each group, and \textbf{Blend}, a continuous heuristic that sorts items by a convex combination of their score-rank and a fair-target rank.
We defer detailed descriptions of all four baselines, including their parameters and implementation choices, to Appendix~\ref{app:baselines}.

All methods are evaluated on the same axes: achieved continuous fairness $\phi$ (computed on the original, undiscretized attribute) and weighted Kemeny distance from the score-based ranking.
On synthetic instances with $n\in\{100,1000,5000\}$ and score--attribute correlation $\mathrm{corr}\in\{0, 0.5, 1\}$, FBS consistently traces a Pareto front that dominates all baselines (full results in Appendix~\ref{app:baselines-extra}).

\begin{figure}[h!]
    \centering
    \includegraphics[width=\textwidth]{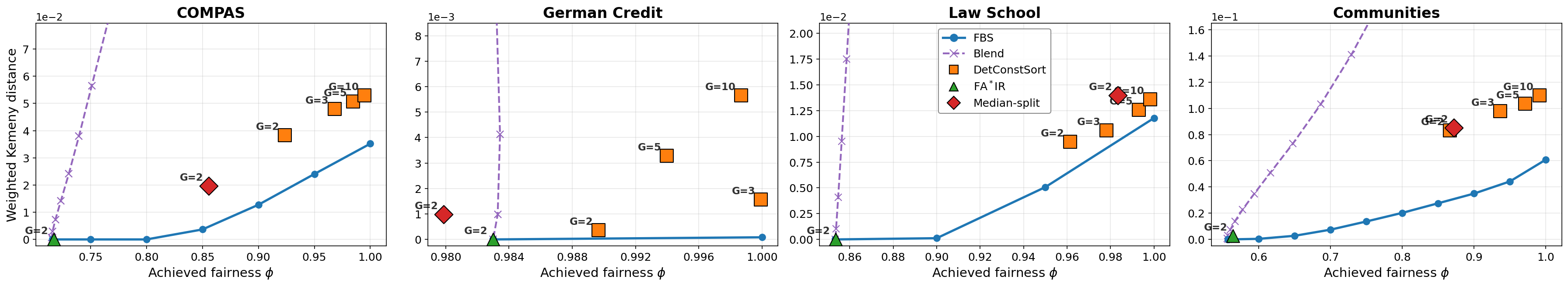}
    \vspace{-4mm}
    \caption{FBS vs.\ discretization baselines on real datasets. FBS (blue line) traces a Pareto front by sweeping $\psi$; baselines, except the continuous heuristic Blend, are shown as individual points for each $G$. FBS dominates all baselines across all four datasets.}
    \label{fig:baselines-real}
\end{figure}

\spara{Real-world datasets.}
We apply FBS to four datasets commonly used in the fair-ranking literature, reporting for each the initial fairness $\phi_0:=\phi(r_s)$ of the score-based ranking: \textbf{COMPAS} ($n=7{,}214$, $\phi_0=0.72$), \textbf{German Credit} ($n=1{,}000$, $\phi_0=0.98$), \textbf{Law School} ($n=18{,}692$, $\phi_0=0.85$), and \textbf{Communities and Crime} ($n=1{,}994$, $\phi_0=0.56$). Detailed descriptions of each dataset (score and attribute choice, source, preprocessing) are deferred to Appendix~\ref{app:datasets}.
For each dataset we sweep the fairness threshold $\psi$ from $0$ to $1$ in steps of $0.05$ and record the achieved fairness $\phi$, the weighted Kemeny distance from the original score-based ranking, and the number of swaps $K^{\ast}$. Results are reported in Figure~\ref{fig:baselines-real} .
On all four datasets, FBS traces a smooth Pareto front that dominates the baselines: DetConstSort with $G\in\{2,3,5,10\}$ and FA*IR ($G=2$) achieve comparable fairness levels but at substantially higher cost, while median-split interleaving reaches moderate fairness at moderate cost but cannot match FBS's efficiency.
The advantage of FBS is most pronounced on COMPAS and Communities, where the initial ranking is far from fair ($\phi_0=0.72$ and $\phi_0=0.56$, respectively) and the continuous attribute carries fine-grained information that is lost by discretization.
German Credit is nearly fair to begin with ($\phi_0=0.98$), so FBS achieves $\phi\approx 1$ at negligible cost.

Appendix~\ref{app:extra-experiments} extends this comparison in three directions. Two repair-then-sort baselines, which first remove the dependence of the score on the attribute and then sort, reach the same fairness as FBS only at a higher cost on all four datasets (Appendix~\ref{app:repair-baselines}). NDCG with the original score as relevance, a utility measure that weighs the top of the list, stays at or above $0.98$ for FBS at the fairness of the best baseline, and a variant of FBS that uses the NDCG loss of a swap as its cost keeps NDCG at $1.00$ on the top $10$ and top $100$ positions at the same fairness (Appendix~\ref{app:ndcg}). Finally, on the three datasets whose initial ranking is far from fair, FBS also reduces the Spearman correlation, the distance correlation, the Hilbert--Schmidt independence criterion and a positional discrepancy between position and attribute as $\psi$ increases (Appendix~\ref{app:dependence}).

\section{Runtime and Scalability}\label{sec:complexity}
We first report theoretical runtime and space complexity of FBS, then the empirical  scalability.
\begin{proposition}[Runtime and space complexity of FBS]\label{prop:complexity}
Let $K^{\ast}=\lceil(\Delta_0-\Delta^{\ast})/2\rceil$ be the number of swaps performed by FBS. Both variants use $O(n)$ auxiliary space beyond the input. The running times are $O(n\log n+K^{\ast})$ (for the unweighted case) and  $O\!\bigl((n+K^{\ast})\log n\bigr)$ (for the weighted case). In both cases the worst case is $K^{\ast}=\Theta(n^{2})$.
%
%
\end{proposition}

\noindent\emph{Proof.} See Appendix~\ref{app:proof-complexity}.

\begin{figure}[t]
    \centering
    \begin{tabular}{cc}
 \hspace{-8mm}\includegraphics[width=.52\textwidth]{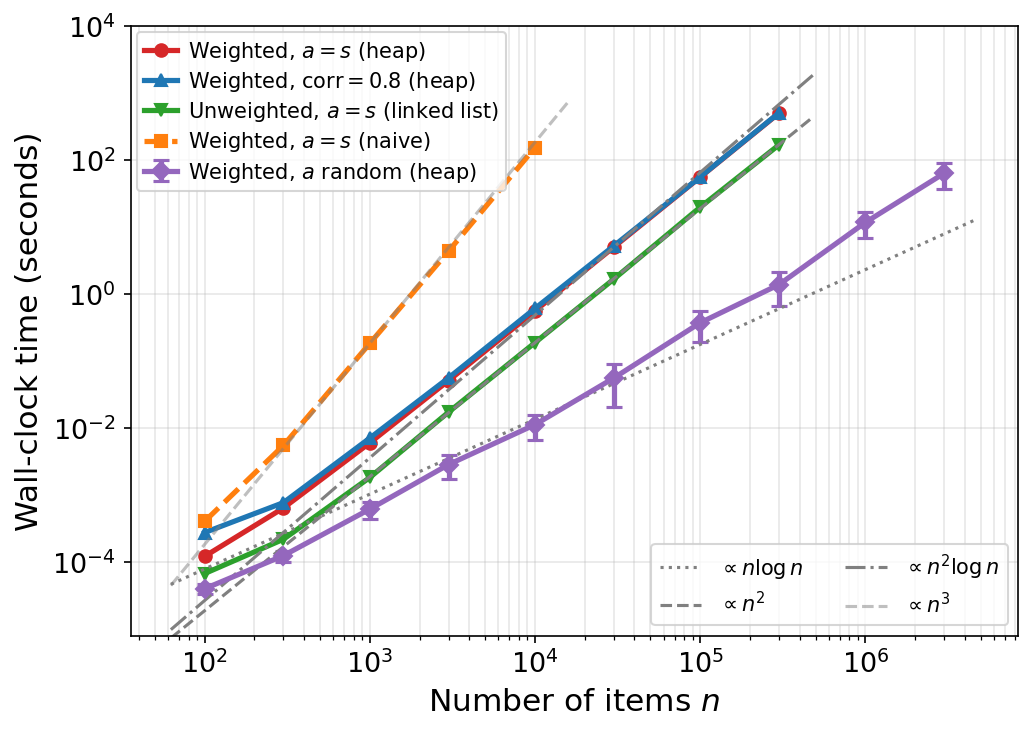} &
  \hspace{-4mm}\includegraphics[width=.52\textwidth]{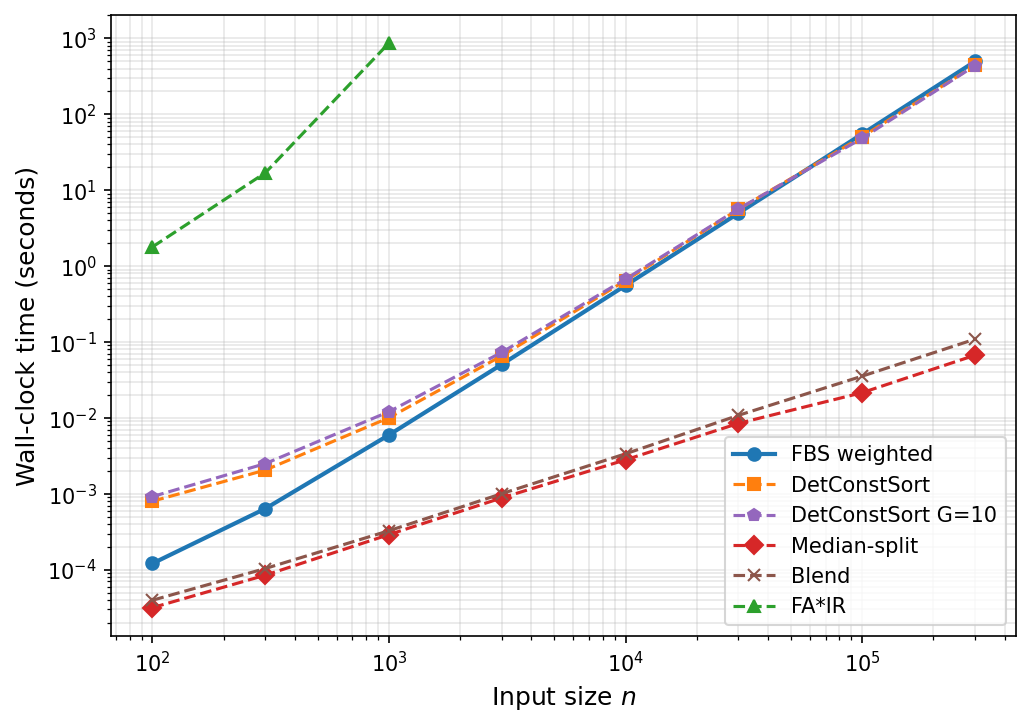}\\
      (a) & (b)
    \end{tabular}
    \vspace{-2mm}
    \caption{(a) Wall-clock time vs.\ input size $n$ (log-log scale). The weighted heap-based FBS scales as $O(n^2\log n)$ in the correlated regime ($a=s$ and $\mathrm{corr}=0.8$) and sub-quadratically when $K^{\ast}$ is small (random attributes). The unweighted linked-list variant scales as $O(n^2)$, while the na\"ive linear-scan implementation scales as $O(n^3)$. (b) Time comparison of FBS vs.\ baselines ($\mathrm{corr}=1$, $\psi=1.0$).}
    \label{fig:scalability}
    \vspace{-4mm}
\end{figure}

\spara{Empirical scalability analysis.} We measure wall-clock time of FBS as $n$ grows from $100$ to $3\times 10^6$.
The following configurations are compared:
\begin{itemize}
    \item \textbf{Weighted heap} (Algorithm~\ref{alg:fbs-weighted}) with $a=s$ (perfectly correlated) and with $\mathrm{corr}(s,a)=0.8$, both at $\psi=1.0$: $K^{\ast}=\Theta(n^2)$, exercising the worst-case $O(n^2\log n)$ branch.
    \item \textbf{Unweighted linked list} (Algorithm~\ref{alg:fbs-unweighted}) with $a=s$, $\psi=1.0$: each swap is $O(1)$, giving $O(K^{\ast})=O(n^2)$ total.
    \item \textbf{Na\"ive linear scan} with $a=s$, $\psi=1.0$: $O(n)$ per swap, giving $O(nK^{\ast})=O(n^3)$ total.
    \item \textbf{Random regime} ($a$ uniform random, $\psi=1.0$): $K^{\ast}$ is much smaller than in the correlated case, yielding sub-quadratic runtime (10 independent instances per~$n$).
\end{itemize}
Figure~\ref{fig:scalability}(a) shows the results on a log-log scale with reference lines for $n\log n$, $n^2$, $n^2\log n$, and $n^3$.
The weighted heap closely tracks $n^2\log n$ in both correlated settings ($a=s$ and $\mathrm{corr}=0.8$), the unweighted linked-list variant tracks $n^2$, and the na\"ive variant follows $n^3$.
In the random regime, $K^{\ast}$ is much smaller and the runtime is sub-quadratic; error bars (one standard deviation over 10 random instances) show moderate variance that increases with~$n$, reflecting the instance-dependent nature of~$K^{\ast}$. A runtime comparison of FBS against the baselines is reported in Figure~\ref{fig:scalability}(b).

\section{Conclusions}\label{sec:conclusions}
We studied the problem of producing a ranking that is as close as possible to a given score-based ranking while  satisfying a fairness constraint defined on a \emph{continuous} protected attribute, 
without resorting to discretization. To this end, we introduced the fairness notion of 
\emph{Continuous Pairwise Parity} and proposed Fair Bubble Sort, 
a greedy adjacent-swap algorithm that minimises the distance from the score-based ranking under given fairness constraint. On the theoretical side, we proved that FBS is \emph{optimal} for the unweighted 
Kemeny distance (Theorem~\ref{thm:unweighted-optimality}) and provided a data-dependent approximation bound for the weighted case. For the case of scores congruent to $[n]$ we sharpened this bound to a  constant $(1{+}\sqrt{2})/2\approx 1.207$ in the worst-case attribute configuration and for $n$ a power of two (Proposition~\ref{prop:ascending-bound}), 
and extensive computation suggests the ratio stays well below~$2$ for arbitrary attributes. On four real datasets (COMPAS, German Credit, Law School, and Communities and Crime) 
FBS dominates the Pareto front of fairness vs.\ accuracy across all tested fairness thresholds, outperforming both 
group-based baselines and a continuous heuristic. 

\paragraph{Limitations.}
Our results have several limitations that suggest natural directions for future work.
$(i)$~The $2$-approximation is conjectured but unproved: while $d_{\mathrm{FBS}}<2\,d^{\ast}$ is empirically validated in every tested configuration, a formal proof remains open.
$(ii)$~FBS handles a \emph{single} continuous protected attribute; extending to multiple attributes simultaneously adds further combinatorial constraints and is left for future work.
$(iii)$~Our setting is \emph{static and one-shot}: long-term dynamics, repeated exposure, and feedback loops~\cite{GiulianiMC024} are beyond the scope of this paper.

\begin{ack}
The authors received no third-party funding for this work and declare no competing interests.
\end{ack}

\bibliographystyle{plain}
\bibliography{biblio}

@misc{fairsearchcore,
  title        = {fairsearchcore: A {P}ython library with the core algorithms used to do fair search},
  author       = {Kitanovski, Ivan},
  year         = {2019},
  howpublished = {Python package, version 1.0.4},
  note         = {\url{https://pypi.org/project/fairsearchcore/}}
}

@inproceedings{GiulianiMC024,
  author       = {Luca Giuliani and
                  Eleonora Misino and
                  Roberta Calegari and
                  Michele Lombardi},
  editor       = {Roberta Calegari and
                  Virginia Dignum and
                  Barry O'Sullivan},
  title        = {Long-Term Fairness Strategies in Ranking with Continuous Sensitive
                  Attributes},
  booktitle    = {Proceedings of the 2nd Workshop on Fairness and Bias in {AI} co-located
                  with 27th European Conference on Artificial Intelligence {(ECAI} 2024),
                  Santiago de Compostela, Spain, October 20th, 2024},
  series       = {{CEUR} Workshop Proceedings},
  publisher    = {CEUR-WS.org},
  year         = {2024},
  url          = {https://ceur-ws.org/Vol-3808/paper11.pdf}
}

@InProceedings{pmlr-v97-mary19a,
  title = 	 {Fairness-Aware Learning for Continuous Attributes and Treatments},
  author =       {Mary, Jeremie and Calauz{\`e}nes, Cl{\'e}ment and Karoui, Noureddine El},
  booktitle = 	 {Proceedings of the 36th International Conference on Machine Learning},
  pages = 	 {4382--4391},
  year = 	 {2019},
  editor = 	 {Chaudhuri, Kamalika and Salakhutdinov, Ruslan},
  volume = 	 {97},
  series = 	 {Proceedings of Machine Learning Research},
  month = 	 {09--15 Jun},
  publisher =    {PMLR},
  url = 	 {https://proceedings.mlr.press/v97/mary19a.html}
}

@inproceedings{Chiappa19,
author = {Chiappa, Silvia},
title = {Path-specific counterfactual fairness},
year = {2019},
publisher = {AAAI Press},
booktitle = {Proceedings of the Thirty-Third AAAI Conference on Artificial Intelligence and Thirty-First Innovative Applications of Artificial Intelligence Conference and Ninth AAAI Symposium on Educational Advances in Artificial Intelligence},
location = {Honolulu, Hawaii, USA},
series = {AAAI'19/IAAI'19/EAAI'19}
}

@inproceedings{kusner2017counterfactual,
 author = {Kusner, Matt J and Loftus, Joshua and Russell, Chris and Silva, Ricardo},
 booktitle = {Advances in Neural Information Processing Systems},
 editor = {I. Guyon and U. Von Luxburg and S. Bengio and H. Wallach and R. Fergus and S. Vishwanathan and R. Garnett},
 pages = {},
 publisher = {Curran Associates, Inc.},
 title = {Counterfactual Fairness},
 url = {https://proceedings.neurips.cc/paper_files/paper/2017/file/a486cd07e4ac3d270571622f4f316ec5-Paper.pdf},
 volume = {30},
 year = {2017}
}

@InProceedings{pmlr-v80-komiyama18a,
  title = 	 {Nonconvex Optimization for Regression with Fairness Constraints},
  author =       {Komiyama, Junpei and Takeda, Akiko and Honda, Junya and Shimao, Hajime},
  booktitle = 	 {Proceedings of the 35th International Conference on Machine Learning},
  pages = 	 {2737--2746},
  year = 	 {2018},
  editor = 	 {Dy, Jennifer and Krause, Andreas},
  volume = 	 {80},
  series = 	 {Proceedings of Machine Learning Research},
  month = 	 {10--15 Jul},
  publisher =    {PMLR},
  url = 	 {https://proceedings.mlr.press/v80/komiyama18a.html}
}

@article{kuhlman2020rank,
  title={Rank aggregation algorithms for fair consensus},
  author={Kuhlman, Caitlin and Rundensteiner, Elke},
  journal={Proceedings of the VLDB Endowment},
  volume={13},
  number={12},
  year={2020}
}

@inproceedings{narasimhan2020pairwise,
  title={Pairwise fairness for ranking and regression},
  author={Narasimhan, Harikrishna and Cotter, Andrew and Gupta, Maya and Wang, Serena},
  booktitle={Proceedings of the AAAI Conference on Artificial Intelligence},
  volume={34},
  pages={5248--5255},
  year={2020}
}

@inproceedings{gretton2005measuring,
  title={Measuring statistical dependence with Hilbert-Schmidt norms},
  author={Gretton, Arthur and Bousquet, Olivier and Smola, Alex and Sch{\"o}lkopf, Bernhard},
  booktitle={International conference on algorithmic learning theory},
  pages={63--77},
  year={2005},
  organization={Springer}
}

@article{zehlike2020matching,
  title={Matching code and law: achieving algorithmic fairness with optimal transport},
  author={Zehlike, Meike and Hacker, Philipp and Wiedemann, Emil},
  journal={Data Mining and Knowledge Discovery},
  volume={34},
  number={1},
  pages={163--200},
  year={2020},
  publisher={Springer},
}

@inproceedings{feldman2015,
author = {Feldman, Michael and Friedler, Sorelle A. and Moeller, John and Scheidegger, Carlos and Venkatasubramanian, Suresh},
title = {Certifying and Removing Disparate Impact},
year = {2015},
booktitle = {Proceedings of the 21th ACM SIGKDD International Conference on Knowledge Discovery and Data Mining},
pages = {259–268},
series = {KDD '15}
}

@inproceedings{HardtPNS16,
  author    = {Moritz Hardt and
               Eric Price and
               Nati Srebro},
  title     = {Equality of Opportunity in Supervised Learning},
  booktitle = {{NeurIPS}},
  pages     = {3315--3323},
  year      = {2016}
}

@inproceedings{2021maxminkdd,
author = {Garc\'{\i}a-Soriano, David and Bonchi, Francesco},
title = {Maxmin-Fair Ranking: Individual Fairness under Group-Fairness Constraints},
year = {2021},
booktitle = {Proceedings of the 27th ACM SIGKDD Conference on Knowledge Discovery and Data Mining},
pages = {436–446},
series = {KDD '21}
}

@inproceedings{YangS17,
  author    = {Ke Yang and
               Julia Stoyanovich},
  title     = {Measuring Fairness in Ranked Outputs},
  booktitle = {{SSDBM}},
  pages     = {22:1--22:6},
  year      = {2017}
}

@inproceedings{CelisSV18,
  author    = {L. Elisa Celis and
               Damian Straszak and
               Nisheeth K. Vishnoi},
  title     = {Ranking with Fairness Constraints},
  booktitle = {{ICALP}},
  year      = {2018}
}

@inproceedings{SinghJ18,
  author    = {Ashudeep Singh and
               Thorsten Joachims},
  title     = {Fairness of Exposure in Rankings},
  booktitle = {{KDD}},
  pages     = {2219--2228},
  year      = {2018}
}

@inproceedings{YangGS19,
  author    = {Ke Yang and
               Vasilis Gkatzelis and
               Julia Stoyanovich},
  title     = {Balanced Ranking with Diversity Constraints},
  booktitle = {{IJCAI}},
  pages     = {6035--6042},
  year      = {2019}
}

@inproceedings{AsudehJS019,
  author    = {Abolfazl Asudeh and
               H. V. Jagadish and
               Julia Stoyanovich and
               Gautam Das},
  title     = {Designing Fair Ranking Schemes},
  booktitle = {{SIGMOD}},
  pages     = {1259--1276},
  year      = {2019}
}

@inproceedings{saito2022fair,
  title={Fair ranking as fair division: Impact-based individual fairness in ranking},
  author={Saito, Yuta and Joachims, Thorsten},
  booktitle={Proceedings of the 28th ACM SIGKDD Conference on Knowledge Discovery and Data Mining},
  pages={1514--1524},
  year={2022}
}

@inproceedings{biega2018equity,
  title={Equity of attention: Amortizing individual fairness in rankings},
  author={Biega, Asia J and Gummadi, Krishna P and Weikum, Gerhard},
  booktitle={The 41st international acm sigir conference on research \& development in information retrieval},
  pages={405--414},
  year={2018}
}

@inproceedings{dworkfairness2012,
author = {Dwork, Cynthia and Hardt, Moritz and Pitassi, Toniann and Reingold, Omer and Zemel, Richard},
title = {Fairness through Awareness},
year = {2012},
booktitle = {Proceedings of the 3rd Innovations in Theoretical Computer Science Conference},
pages = {214–226},
series = {ITCS '12}
}

@inproceedings{fairrankingcrticial2022,
author = {Patro, Gourab K. and Porcaro, Lorenzo and Mitchell, Laura and Zhang, Qiuyue and Zehlike, Meike and Garg, Nikhil},
title = {Fair Ranking: A Critical Review, Challenges, and Future Directions},
year = {2022},
booktitle = {2022 ACM Conference on Fairness, Accountability, and Transparency},
pages = {1929–1942},
series = {FAccT '22}
}

@article{algfairnessdef2021,
author = {Mitchell, Shira and Potash, Eric and Barocas, Solon and D'Amour, Alexander and Lum, Kristian},
title = {Algorithmic Fairness: Choices, Assumptions, and Definitions},
journal = {Annual Review of Statistics and Its Application},
volume = {8},
number = {1},
pages = {141-163},
year = {2021}
}

@inproceedings{Jag19b,
author = {Jagadish, H. V. and Bonchi, Francesco and Eliassi-Rad, Tina and Getoor, Lise and Gummadi, Krishna and Stoyanovich, Julia},
title = {The Responsibility Challenge for Data},
year = {2019},
booktitle = {Proceedings of the 2019 International Conference on Management of Data},
pages = {412–414},
location = {Amsterdam, Netherlands},
series = {SIGMOD '19}
}

@article{pitoura2022fairness,
  title={Fairness in rankings and recommendations: an overview},
  author={Pitoura, Evaggelia and Stefanidis, Kostas and Koutrika, Georgia},
  journal={The VLDB Journal},
  volume={31},
  number={3},
  pages={431--458},
  year={2022},
  publisher={Springer}
}

@article{zehlike2022fairness,
  title={Fairness in ranking, part i: Score-based ranking},
  author={Zehlike, Meike and Yang, Ke and Stoyanovich, Julia},
  journal={ACM Computing Surveys},
  volume={55},
  number={6},
  pages={1--36},
  year={2022},
  publisher={ACM New York, NY}
}

@article{szekely2007measuring,
  title   = {Measuring and testing dependence by correlation of distances},
  author  = {Sz{\'e}kely, G{\'a}bor J and Rizzo, Maria L and Bakirov, Nail K},
  journal = {Annals of Statistics},
  volume  = {35},
  number  = {6},
  pages   = {2769--2794},
  year    = {2007}
}

@inproceedings{Zehlike2017,
  title     = {FA*IR: A fair top-k ranking algorithm},
  author    = {Zehlike, Meike and Bonchi, Francesco and Castillo, Carlos and Hajian, Sara and Megahed, Mohamed and Baeza-Yates, Ricardo},
  booktitle = {Proceedings of the 2017 ACM Conference on Information and Knowledge Management (CIKM)},
  pages     = {1569--1578},
  year      = {2017}
}

@article{angwin2016machine,
  author = {Angwin, Julia and Larson, Jeff and Mattu, Surya and Kirchner, Lauren},
  title = {Machine Bias: There's Software Used Across the Country to Predict Future Criminals. And It's Biased Against Blacks},
  journal = {ProPublica},
  year = {2016},
  month = {May},
  url = {https://www.propublica.org/article/machine-bias-risk-assessments-in-criminal-sentencing}
}

@book{knuth1998art,
  title={The Art of Computer Programming: Sorting and Searching, volume 3},
  author={Knuth, Donald E},
  year={1998},
  publisher={Addison-Wesley Professional}
}

@article{diaconis1977spearman,
  title={Spearman's footrule as a measure of disarray},
  author={Diaconis, Persi and Graham, Ronald L},
  journal={Journal of the Royal Statistical Society Series B: Statistical Methodology},
  volume={39},
  number={2},
  pages={262--268},
  year={1977},
  publisher={Oxford University Press}
}

@inproceedings{geyik2019fairness,
  title={Fairness-aware ranking in search \& recommendation systems with application to linkedin talent search},
  author={Geyik, Sahin Cem and Ambler, Stuart and Kenthapadi, Krishnaram},
  booktitle={Proceedings of the 25th acm sigkdd international conference on knowledge discovery \& data mining},
  pages={2221--2231},
  year={2019}
}

@inproceedings{cachel2024fairranktune,
  title={Fairranktune: A python toolkit for fair ranking tasks},
  author={Cachel, Kathleen and Rundensteiner, Elke},
  booktitle={Proceedings of the 33rd ACM International Conference on Information and Knowledge Management},
  pages={5195--5199},
  year={2024}
}

@misc{statlog,
  author       = {Hofmann, Hans},
  title        = {{Statlog (German Credit Data)}},
  year         = {1994},
  howpublished = {UCI Machine Learning Repository},
  note         = {{DOI}: https://doi.org/10.24432/C5NC77}
}

@misc{communities_and_crime_183,
  author       = {Redmond, Michael},
  title        = {{Communities and Crime}},
  year         = {2002},
  howpublished = {UCI Machine Learning Repository},
  note         = {{DOI}: https://doi.org/10.24432/C53W3X}
}

@techreport{wightman1998lsac,
  title       = {{LSAC} National Longitudinal Bar Passage Study},
  author      = {Wightman, Linda F.},
  institution = {Law School Admission Council},
  type        = {{LSAC} Research Report Series},
  year        = {1998}
}

@article{kelly2000inequality,
  title={Inequality and crime},
  author={Kelly, Morgan},
  journal={Review of economics and Statistics},
  volume={82},
  number={4},
  pages={530--539},
  year={2000},
  publisher={MIT Press 238 Main St., Suite 500, Cambridge, MA 02142-1046, USA journals~…}
}

@article{johndrow2019algorithm,
  title={An algorithm for removing sensitive information},
  author={Johndrow, James E and Lum, Kristian},
  journal={The Annals of Applied Statistics},
  volume={13},
  number={1},
  pages={189--220},
  year={2019},
  publisher={JSTOR}
}

@inproceedings{fagin2004comparing,
  title={Comparing and aggregating rankings with ties},
  author={Fagin, Ronald and Kumar, Ravi and Mahdian, Mohammad and Sivakumar, D and Vee, Erik},
  booktitle={Proceedings of the twenty-third ACM SIGMOD-SIGACT-SIGART symposium on Principles of database systems},
  pages={47--58},
  year={2004}
}

@inproceedings{conitzer2006improved,
  title={Improved bounds for computing Kemeny rankings},
  author={Conitzer, Vincent and Davenport, Andrew and Kalagnanam, Jayant},
  booktitle={AAAI},
  volume={6},
  pages={620--626},
  year={2006}
}

@incollection{kemeny1962preference,
  title     = {Preference ranking: An axiomatic approach},
  author    = {Kemeny, John G. and Snell, J. Laurie},
  booktitle = {Mathematical Models in the Social Sciences},
  publisher = {Ginn},
  address   = {Boston, MA, USA},
  pages     = {9--23},
  year      = {1962}
}

\clearpage
\appendix


\begin{center}
    \LARGE\textbf{Technical Appendix and Supplementary Material}
\end{center}
\vspace{1cm}

\begin{center}
    \Large\textbf{Appendix Contents}
\end{center}

\startcontents[appendices]
\printcontents[appendices]{l}{1}{\setcounter{tocdepth}{2}}

\clearpage


\section{Broader Impact}\label{sec:broader-impact}

Fair ranking algorithms are deployed in high-stakes settings including hiring, lending, college admissions, content moderation, and criminal justice. Almost all existing algorithms require the protected attribute to be discrete, which forces practitioners to either categorise inherently continuous variables (age, income, health scores) into ad-hoc bins or to abandon fairness control entirely. Both choices are problematic: the former discards information and introduces arbitrary thresholds that can themselves be unfair (e.g., who decides where the age bins begin and end?), while the latter offers no protection at all. By providing a theoretically grounded algorithm that operates directly on continuous attributes, our work lowers the barrier to using fair-ranking methods in these settings and removes a class of arbitrary modelling choices that have historically been difficult to justify.

At the same time, we caution against treating FBS---or any post-processing fairness intervention---as a complete solution. The fairness measure $\phi$ targets pairwise statistical parity with respect to a continuous attribute, but it does not by itself guarantee proportional top-$k$ representation, equal exposure, calibration across sub-populations, or freedom from indirect discrimination through correlated features. Rankings that look fair under $\phi$ can still encode biases inherited from the underlying score function, and a $\phi$-fair re-ranking may give a misleading impression of fairness when applied on top of a flawed score. Responsible use therefore requires (a)~auditing the score function itself, (b)~verifying that $\phi$ is the appropriate fairness criterion for the application context, and (c)~combining FBS with complementary checks (e.g., exposure or representation audits) where multiple fairness notions are simultaneously relevant. The choice of which protected attribute to enforce fairness on is itself a normative decision that should involve the affected stakeholders rather than be made unilaterally by the system designer.

\section{Motivations for the Continuous Pairwise Parity (CPP) measure $\phi$}\label{app:motivations}

This section motivates the choice of $\phi(r)=1-|\tau(r,a)|$ as the fairness measure for ranking with continuous sensitive attributes. We argue that (i)~$\phi$ is the natural pairwise extension of statistical parity to the continuous-attribute setting; (ii)~it strictly generalises the Pairwise Statistical Parity of~\cite{kuhlman2020rank} from discrete groups to continuous values; and (iii)~it satisfies a list of desirable properties---boundedness, symmetry, discretisation-freeness, and tractability---that the standard group-based measures cannot meet without arbitrary binning.

\paragraph{From group statistical parity to a pairwise principle.}
In classification, \emph{statistical parity}~\cite{dworkfairness2012} requires that the probability of a positive outcome be equal across protected groups, $\Pr(\hat{y}=1\mid g=A)=\Pr(\hat{y}=1\mid g=B)$. Adapting this principle to ranking is non-trivial: a ``positive outcome'' has no immediate analogue in a permutation, so the literature replaces it with proxies such as ``appearing in the top-$k$''~\cite{Zehlike2017,CelisSV18,SinghJ18}, which require predefined groups~$A,B$ and a discrete cut-off~$k$. When the protected attribute is continuous, neither the groups nor the cut-off are defined: discretising~$a$ into bins forces an arbitrary choice of thresholds, discards fine-grained variation, and destroys the ordinality of the attribute (e.g.\ the bin $[20,30)$ is treated as categorically distinct from $[30,40)$, even though their boundary is conventional).

A more natural primitive in the ranking setting is the \emph{pair}: for any two items~$u,v$, one is ranked above the other, and one has a larger attribute value than the other. Pairwise statistical parity~\cite{kuhlman2020rank} formalises fairness as the requirement that the conditional probability of one group being ranked above the other equals the converse:
\[
\Pr\!\bigl(r(u)<r(v)\mid g(u)=A,\,g(v)=B\bigr)
=\Pr\!\bigl(r(u)<r(v)\mid g(u)=B,\,g(v)=A\bigr).
\]
The translation to a continuous attribute is direct: replace ``$g(u)=A,\,g(v)=B$'' by ``$a(u)>a(v)$'' (and the converse by ``$a(u)<a(v)$''), so that fairness becomes the requirement that, for a uniformly random pair $(u,v)$, the higher-attribute item is equally likely to be ranked above or below the lower-attribute one. Empirically over the $n$ items, this reads $N_{\nearrow}(r)\approx N_{\searrow}(r)$, which is precisely the condition $\tau(r,a)\approx 0$, equivalently $\phi(r)\approx 1$.

\paragraph{Reduction to Pairwise Statistical Parity in the binary case.}
The CPP measure recovers, as a special case, the Pairwise Statistical Parity of~\cite{kuhlman2020rank}, a notion also defined by~\cite{narasimhan2020pairwise}. Suppose $a$ takes only two values, $a(u)\in\{0,1\}$, encoding membership in a low group $A=\{u:a(u)=0\}$ and a high group $B=\{u:a(u)=1\}$ of sizes $n_A$ and $n_B$; this is the situation obtained by discretising a continuous attribute at a threshold, for example its median. Let
\[
P(r):=\Pr\bigl(r(v)<r(u)\mid u\in A,\;v\in B\bigr)
\]
be the probability that a uniformly random $B$-item is ranked above a uniformly random $A$-item. The rank-parity score of~\cite{kuhlman2020rank} is $P(r)$ for group $B$ and $1-P(r)$ for group $A$, and their Pairwise Statistical Parity requires the two scores to be equal, i.e.\ $P(r)=\tfrac12$.

\begin{proposition}[Binary CPP equals Pairwise Statistical Parity]\label{prop:binary-parity}
For a binary attribute as above, $\Delta(r)=n_A n_B\,|2P(r)-1|$, and hence
\[
\phi(r)=1-\frac{n_A n_B}{\binom{n}{2}}\,\bigl|2P(r)-1\bigr|.
\]
In particular, $\phi(r)=1$ if and only if $P(r)=\tfrac12$, i.e.\ if and only if $r$ satisfies the Pairwise Statistical Parity of~\cite{kuhlman2020rank}.
\end{proposition}

\begin{proof}
Two items of the same group have equal attribute values, so their pair contributes to neither $N_{\nearrow}$ nor $N_{\searrow}$ (tie convention of Section~\ref{sec:preliminaries}); only the $n_A n_B$ cross-group pairs count. Among them, $n_A n_B P(r)$ pairs have the $B$-item ranked above the $A$-item and $n_A n_B(1-P(r))$ have it ranked below. By the definition of Section~\ref{sec:preliminaries}, a cross-group pair is concordant when the item with the larger attribute value, the $B$-item, has the larger position index, i.e.\ is ranked below the $A$-item; hence $N_{\nearrow}=n_A n_B(1-P(r))$, $N_{\searrow}=n_A n_B P(r)$, and $\Delta(r)=|N_{\nearrow}-N_{\searrow}|=n_A n_B|2P(r)-1|$. Dividing by $\binom{n}{2}$ gives $|\tau(r,a)|$ and the expression for $\phi(r)$.
\end{proof}

The quantity $P(r)$ is the cross-group AUC, i.e.\ the Mann--Whitney $U$ statistic of the two groups' positions divided by $n_A n_B$, and $2P(r)-1$ is Somers' $D$. CPP therefore subsumes the discrete-group definition without loss of expressivity, and discretising a continuous attribute recovers that metric exactly. For $G>2$ ordered groups the same computation gives $N_{\nearrow}-N_{\searrow}=\sum_{g<h} n_g n_h\,(1-2P_{gh}(r))$, where $P_{gh}(r)$ is the probability that a random item of the higher group $h$ is ranked above a random item of the lower group $g$: the signed imbalance is the size-weighted sum of the pairwise parity terms over all pairs of groups.

\paragraph{Statistical interpretation.}
At the strict-fairness threshold $\psi=1$ (equivalently $\Delta^{\ast}=0$ when $n(n-1)/2$ is even), $\phi(r)=1$ admits a probabilistic reading: for a uniformly random ordered pair of distinct items~$(u,v)$,
\[
\Pr_{(u,v)}\!\bigl(r(u)<r(v)\mid a(u)>a(v)\bigr)=\tfrac{1}{2},
\]
i.e.\ the relative ranking position carries no monotonic signal about the attribute. Intermediate thresholds $\psi\in(0,1)$ correspond to a controlled relaxation of this equality, with the maximum imbalance $\Delta^{\ast}=\lfloor n(n-1)(1-\psi)/2\rfloor$ measuring how far from the strict statistical-parity condition the practitioner is willing to deviate.

\paragraph{Comparison with related pairwise notions.}
\cite{narasimhan2020pairwise} introduced a pairwise fairness criterion for learning-to-rank and regression, requiring equality of the probability that a relevant item from group~$A$ is ranked above a non-relevant item from group~$B$, and vice versa. Their formulation targets the supervised-learning setting (training a scoring function under a fairness constraint) and relies on discrete groups. CPP, in contrast, is defined directly on the output ranking and on a continuous attribute, making it suitable for post-processing without retraining. The two notions agree in spirit---both anchor fairness at the pairwise level---but operate in different regimes.

\paragraph{Properties of $\phi$.}
The CPP measure satisfies the following desiderata, none of which are jointly achieved by the standard group-based ranking-fairness measures.
\begin{enumerate}
\item \textbf{Boundedness and calibration.} $\phi(r)\in[0,1]$ for every ranking, with $\phi(r)=1$ iff $\tau(r,a)=0$ (perfect fairness) and $\phi(r)=0$ iff $|\tau(r,a)|=1$ (deterministic monotonic dependence between ranking and attribute).
\item \textbf{Direction symmetry.} $\phi$ is invariant under reflection of the attribute, $\phi(r;a)=\phi(r;-a)$: the measure does not privilege ``high-attribute on top'' over ``low-attribute on top''.
\item \textbf{Discretisation-freeness.} $\phi$ is defined directly on the continuous values $a(u)$ and requires neither thresholds nor groups; in particular, it is invariant under any strictly monotone transformation of~$a$.
\item \textbf{Robustness to attribute perturbations.} A small perturbation of the attribute values that preserves their relative order leaves $\phi$ exactly unchanged; an attribute swap that flips a single pair changes $\phi$ by~$O(1/n^2)$. CPP is therefore stable under measurement noise that preserves rank order.
\item \textbf{Pairwise locality.} $\phi$ decomposes additively over pairs, mirroring the pairwise structure of the (unweighted) Kemeny distance. This shared structure is what enables FBS to optimise both quantities through a single algorithmic primitive---the adjacent swap---and underlies the optimality result of Theorem~\ref{thm:unweighted-optimality}.
\item \textbf{Tractability.} $\phi(r)$ is computable in $O(n\log n)$ via merge-sort--based inversion counting, matching the cost of evaluating the Kemeny distance.
\end{enumerate}

\paragraph{Comparison with alternative dependence measures.}
Other measures of dependence between position and attribute could replace the Kendall-$\tau$ correlation in the definition of $\phi$. Spearman's correlation is also monotone, but it measures squared rank differences rather than pairwise concordance, so it does not share the pairwise structure of the Kemeny distance that lets a single adjacent-swap primitive optimise fairness and utility together. Distance correlation~\cite{szekely2007measuring} and the Hilbert--Schmidt independence criterion (HSIC)~\cite{gretton2005measuring} also capture non-monotone dependence, but they are not additive over pairs, and the effect of an adjacent swap on them depends on the whole ranking, so they admit neither the exact analysis of FBS nor a tractable hard constraint. Binning the attribute and applying a group-parity measure discards its ordinal structure and introduces binning artifacts. The Kendall-based CPP is the notion that is at the same time binning-free, ordinal and pairwise-decomposable, which is what lets one adjacent-swap primitive optimise both fairness and utility; the price is that CPP captures monotone dependence only, as discussed next.

\paragraph{Limitations.}
CPP captures \emph{monotonic} dependence between ranking and attribute. A ranking in which the highest- and lowest-attribute items occupy the middle positions and mid-attribute items are placed at the extremes can have $\tau(r,a)=0$ yet exhibit a clearly non-uniform attribute distribution along positions. Detecting such non-monotonic biases requires complementary measures (e.g., position-conditional moments of~$a$), which we view as orthogonal extensions rather than as substitutes for a pairwise primitive.
\section{Pseudocode of FBS for the weighted Kemeny distance}\label{app:weighted-alg}

When the Kemeny distance is weighted by score gaps $w_{ij}=|s(i)-s(j)|$, different swaps have different costs and FBS must select the cheapest majority-sign adjacent pair at each step.
We replace the doubly-linked list~$L$ of Algorithm~\ref{alg:fbs-unweighted} with a min-heap~$H$ keyed by the adjacent cost $c[k]=|S[k]-S[k+1]|$ (the normalisation $1/W$ where $W=\sum_{k<l}|s(k)-s(l)|$ is a monotone rescaling and plays no role in the arg-min), supporting extraction, insertion, deletion, and key updates in $O(\log n)$.
Ties in cost are broken by picking the leftmost pair (via the secondary key~$k$).

\begin{algorithm}[h!]
\caption{FBS (Fair Bubble Sort) -- weighted, with priority queue}\label{alg:fbs-weighted}
\begin{algorithmic}[1]
\Require Items $U = [n]$, attribute $a: U \to \mathbb{R}_+$, score $s: U \to \mathbb{R}_+$, fairness threshold $\psi\in[0,1]$
\Ensure Ranking $R$ solving Problem~\eqref{eq:optimization_problem} with the weighted Kemeny distance
\State $(R,S,A)\gets$ arrays initialized according to $r_s$ (descending order of score) \Comment{initial $r_s$}
\State $c[k]\gets |S[k]-S[k+1]|$ for $k=1,\ldots,n-1$
\State $\sigma[k]\gets\mathrm{sign}(A[k+1]-A[k])$ for $k=1,\ldots,n-1$ \Comment{adjacent-pair signs}
\State $N_{\nearrow}\gets |\{i<j:A[i]<A[j]\}|$;\quad $N_{\searrow}\gets |\{i<j:A[i]>A[j]\}|$
\State $\Delta^{\ast}\gets\max\!\bigl(\lfloor(1-\psi)\,n(n-1)/2\rfloor,\;|N_{\nearrow}-N_{\searrow}|\!\bmod 2\bigr)$ \Comment{parity-safe threshold}
\State $\mu\gets\mathrm{sign}(N_{\nearrow}-N_{\searrow})$ \Comment{majority sign ($+1$ or $-1$)}
\State $H\gets$ min-heap keyed by $(c[k],\,k)$ over $\{k:\sigma[k]=\mu\}$ \Comment{build in $O(n)$}
\While{$|N_{\nearrow}-N_{\searrow}|>\Delta^{\ast}$}
    \State $k^{\ast}\gets\textsc{Extract-Min}(H)$ \Comment{$O(\log n)$}
    \State swap $R[k^{\ast}]\leftrightarrow R[k^{\ast}+1]$, $S[k^{\ast}]\leftrightarrow S[k^{\ast}+1]$, $A[k^{\ast}]\leftrightarrow A[k^{\ast}+1]$
    \For{$\ell\in\{k^{\ast}-1,\;k^{\ast},\;k^{\ast}+1\}\cap\{1,\ldots,n-1\}$} \Comment{update neighbours}
        \State $c[\ell]\gets|S[\ell]-S[\ell+1]|$
        \State $\sigma_{\mathrm{old}}\gets\sigma[\ell]$
        \If{$\ell=k^{\ast}$}
            \State $\sigma[\ell]\gets-\sigma[\ell]$ \Comment{middle sign flips (Observation~\ref{obs:single-swap})}
        \Else
            \State $\sigma[\ell]\gets\mathrm{sign}(A[\ell+1]-A[\ell])$ \Comment{recompute from new attributes}
        \EndIf
        \If{$\sigma_{\mathrm{old}}=\mu$ \textbf{and} $\sigma[\ell]\ne\mu$}
            \State remove $\ell$ from $H$ \Comment{$O(\log n)$}
        \ElsIf{$\sigma_{\mathrm{old}}\ne\mu$ \textbf{and} $\sigma[\ell]=\mu$}
            \State insert $\ell$ into $H$ with key $(c[\ell],\,\ell)$ \Comment{$O(\log n)$}
        \ElsIf{$\sigma[\ell]=\mu$ \textbf{and} $c[\ell]$ changed}
            \State update key of $\ell$ in $H$ to $(c[\ell],\,\ell)$ \Comment{$O(\log n)$}
        \EndIf
    \EndFor
    \State $N_{\nearrow}\gets N_{\nearrow}-\mu$;\quad $N_{\searrow}\gets N_{\searrow}+\mu$
\EndWhile
\State \Return $R$
\end{algorithmic}
\end{algorithm}

\section{Proofs of Section 5}\label{app:proofs}

\subsection{Proofs of structural lemmas}

\begin{proof}[Proof of Observation~\ref{obs:single-swap} (Single-swap increment)]
Let $u,v\in U$ be the items with $r(u)=k$ and $r(v)=k{+}1$, so that after the adjacent swap $r'(u)=k{+}1$, $r'(v)=k$, and $r'(w)=r(w)$ for every $w\notin\{u,v\}$.

\emph{Pairwise locality.} Let $w$ be any item distinct from $u$ and $v$, with rank $r(w)\notin\{k,k{+}1\}$. Since $r(w)$ is either strictly less than both $k$ and $k{+}1$ or strictly greater than both, its order relative to $u$ (and to $v$) is unchanged by the swap, and the rank of $w$ itself is fixed. Hence for every pair of items other than $\{u,v\}$ the relative order in the ranking is the same in $r$ and $r'$; consequently both the $r_s$-inversion status (which depends only on the relative orders in $r$ and $r_s$) and the Kendall orientation (which depends only on the relative order in $r$ and the attribute values) are preserved. The only pair whose status can change is $\{u,v\}$, whose relative order is reversed by the swap.

\emph{Effect on $d$.} The pair $\{u,v\}$ contributes $w_{uv}$ to $d(r_s,\cdot)$ iff it lies in $\mathrm{Inv}(r_s,\cdot)$, and the swap toggles this membership:
\begin{itemize}\setlength{\itemsep}{1pt}
\item if $\{u,v\}\notin\mathrm{Inv}(r_s,r)$ (i.e.\ $u,v$ are in the same relative order as in $r_s$), then $\{u,v\}\in\mathrm{Inv}(r_s,r')$ and the swap \emph{creates} an inversion, so $d(r_s,r')-d(r_s,r)=+w_{uv}$;
\item if $\{u,v\}\in\mathrm{Inv}(r_s,r)$, then $\{u,v\}\notin\mathrm{Inv}(r_s,r')$ and the swap \emph{resolves} an inversion, so $d(r_s,r')-d(r_s,r)=-w_{uv}$.
\end{itemize}
By pairwise locality no other pair contributes to the change, establishing the first claim.

\emph{Effect on $N_{\nearrow}-N_{\searrow}$.} Under the general-position assumption (no ties in $a$), every pair is either $\nearrow$ or $\searrow$. By pairwise locality the orientation of every pair other than $\{u,v\}$ is preserved, while the orientation of $\{u,v\}$ flips between $\nearrow$ and $\searrow$. Hence $N_{\nearrow}$ changes by exactly $\pm 1$ and $N_{\searrow}$ by exactly $\mp 1$, giving
\[
\bigl(N_{\nearrow}-N_{\searrow}\bigr)(r')-\bigl(N_{\nearrow}-N_{\searrow}\bigr)(r)\in\{-2,+2\},
\]
with sign $+2$ if $\{u,v\}$ was $\searrow$ in $r$ (and becomes $\nearrow$ in $r'$) and $-2$ otherwise.
\end{proof}

\begin{proof}[Proof of Lemma~\ref{lem:contiguous} (Reduction to adjacent swaps)]
Since $r_s$ and $r$ are bijections $U\to[n]$, the map $\pi:=r\circ r_s^{-1}$ is a permutation of $[n]$, and $r=\pi\circ r_s$. It is classical that $S_n$ is generated by the adjacent transpositions $\tau_k:=(k,k{+}1)$, $k=1,\ldots,n{-}1$: every $\pi\in S_n$ admits a finite decomposition $\pi=\tau_{k_m}\circ\tau_{k_{m-1}}\circ\cdots\circ\tau_{k_1}$, and the minimum number of factors equals $|\mathrm{Inv}(\pi)|=|\mathrm{Inv}(r_s,r)|$ (e.g.\ the bound is attained by bubble sort)~\cite{knuth1998art}. An adjacent swap at rank $k$ sends a ranking $r'$ to $\tau_k\circ r'$, so applying $\tau_{k_1},\tau_{k_2},\ldots,\tau_{k_m}$ in order to $r_s$ produces the trajectory
\[
r_s,\ \tau_{k_1}\circ r_s,\ \tau_{k_2}\circ\tau_{k_1}\circ r_s,\ \ldots,\ \tau_{k_m}\circ\cdots\circ\tau_{k_1}\circ r_s=\pi\circ r_s=r,
\]
i.e., an algorithm in $A_{swap}$ whose output is $r$.
\end{proof}

\begin{proof}[Proof of Lemma~\ref{lem:nondecreasing} (Reduction to non-decreasing swaps)]
We construct a sequence of adjacent swaps from $r_s$ to $r^{\ast}$ along which $d(r_s,\cdot)$ is non-decreasing. Run bubble sort on $r^{\ast}$, sorting items toward the order prescribed by $r_s$: at each step, with current ranking $q$, find adjacent ranks $k, k{+}1$ such that the items $u,v$ at those ranks in $q$ (i.e., $q(u)=k$, $q(v)=k{+}1$) are in opposite $r_s$-order — equivalently, $\{u,v\}$ contributes an inversion to $\mathrm{Inv}(r_s,q)$ — and swap them. Such a pair exists whenever $q\ne r_s$, and each swap reduces $|\mathrm{Inv}(r_s,q)|$ by exactly~$1$, so the algorithm terminates at $r_s$ after $m=|\mathrm{Inv}(r_s,r^{\ast})|$ swaps. By Observation~\ref{obs:single-swap} each swap resolves the $r_s$-inversion of the corresponding item pair $\{u,v\}$ and decreases $d(r_s,\cdot)$ by $w_{uv}\ge 0$; hence the trajectory
\[
r^{\ast}=q^{(0)}\xrightarrow{\tau_{k_1}}q^{(1)}\xrightarrow{\tau_{k_2}}\cdots\xrightarrow{\tau_{k_m}}q^{(m)}=r_s
\]
satisfies $d(r_s,q^{(0)})\ge d(r_s,q^{(1)})\ge\cdots\ge d(r_s,q^{(m)})=0$.

Adjacent transpositions are involutions ($\tau_k^2=\mathrm{id}$), so applying $\tau_{k_m},\tau_{k_{m-1}},\ldots,\tau_{k_1}$ in this reversed order to $r_s$ visits the same rankings in reverse,
\[
r_s=q^{(m)}\to q^{(m-1)}\to\cdots\to q^{(0)}=r^{\ast},
\]
along which $d(r_s,\cdot)$ is non-decreasing. The corresponding algorithm therefore lies in $A_{swap,\nearrow}$ and produces $r^{\ast}$.
\end{proof}

\begin{proof}[Proof of Lemma~\ref{lem:valid-swap-exists} (Existence of a valid greedy swap)]
We argue by contrapositive. The claim ``$N_{\nearrow}(r)>N_{\searrow}(r)\Rightarrow$ some adjacent pair is $\nearrow$'' is logically equivalent to
\[
\text{no adjacent pair of $r$ is $\nearrow$}\ \Longrightarrow\ N_{\nearrow}(r)\le N_{\searrow}(r).
\]
Because the values $b_1,\ldots,b_n$ are pairwise distinct, every pair is either $\nearrow$ or $\searrow$, so the hypothesis ``no adjacent $\nearrow$'' is equivalent to
\[
b_k>b_{k+1}\qquad\text{for every }k=1,\ldots,n-1,\qquad(\star)
\]
i.e.\ every adjacent pair is $\searrow$. We shall actually establish the \emph{stronger} conclusion $N_{\nearrow}(r)=0$ from $(\star)$; since $N_{\nearrow}(r)=0\le N_{\searrow}(r)$ this suffices.

Assume $(\star)$. For any pair $i<j$, chaining the $j-i$ adjacent strict inequalities in $(\star)$ and using transitivity of $>$ on the distinct reals $b_1,\ldots,b_n$ yields
\[
b_i>b_{i+1}>\cdots>b_j\ \Longrightarrow\ b_i>b_j,
\]
so $(i,j)$ is $\searrow$. This holds for every $i<j$, hence $N_{\nearrow}(r)=0$ as announced. Returning to the original direction: if $N_{\nearrow}(r)>N_{\searrow}(r)\ge 0$ then $N_{\nearrow}(r)\ge 1$, so $(\star)$ must fail, i.e.\ at least one adjacent pair is $\nearrow$. The $\searrow$-majority case is obtained by exchanging $>$ and $<$ throughout.

The argument only invokes transitivity on the pairwise distinct values $b_1,\ldots,b_n$ of the current ranking, so it applies unchanged at every greedy iteration, not only at the initial $r=r_s$.
\end{proof}

\begin{proof}[Proof of Proposition~\ref{prop:minimal-swaps} (Minimality of the number of greedy swaps)]
\emph{Lower bound.} By Observation~\ref{obs:single-swap}, any adjacent swap changes $N_{\nearrow}-N_{\searrow}$ by exactly $\pm 2$. Since $\Delta=|N_{\nearrow}-N_{\searrow}|$, the reverse triangle inequality gives $\Delta'\ge\Delta-2$ after any single swap ($\Delta'$ can equal $\Delta-2$, $\Delta$, or $\Delta+2$; the middle case occurs only when $\Delta=1$ and the swap flips the sign of $N_{\nearrow}-N_{\searrow}$). Going from $\Delta_0$ to any value $\le\Delta^{\ast}$ therefore requires at least $\lceil(\Delta_0-\Delta^{\ast})/2\rceil=K^{\ast}$ swaps.

\emph{Parity remark.} Each swap preserves the parity of $N_{\nearrow}-N_{\searrow}$, so $\Delta\bmod 2=\Delta_0\bmod 2$ throughout. If $\Delta^{\ast}<\Delta_0\bmod 2$, the target $\Delta\le\Delta^{\ast}$ is unreachable. This is precisely why the algorithm sets $\Delta^{\ast}\gets\max(\Delta^{\ast},\,\Delta_0\bmod 2)$: the adjusted threshold ensures $K^{\ast}$ is a non-negative integer and $\Delta_0-2K^{\ast}\ge 0$.

\emph{FBS attains the bound.} We assume without loss of generality $N_{\nearrow}(r_s)\ge N_{\searrow}(r_s)$, so the majority sign is $\mu=+1$ and $N_{\nearrow}-N_{\searrow}=\Delta_0>0$ (the $\mu=-1$ case is symmetric). We prove by induction on~$k$ the loop invariant
\[
N_{\nearrow}-N_{\searrow}=\Delta_0-2k\qquad(0\le k\le K^{\ast}).
\]

\emph{Base case} ($k=0$): immediate from the definition of~$\Delta_0$.

\emph{Inductive step} ($k\to k{+}1$, for $k<K^{\ast}$): by the inductive hypothesis $N_{\nearrow}-N_{\searrow}=\Delta_0-2k$. Since $k<K^{\ast}=\lceil(\Delta_0-\Delta^{\ast})/2\rceil$, we have $\Delta_0-2k>\Delta^{\ast}\ge 0$, so in particular $N_{\nearrow}>N_{\searrow}$. Lemma~\ref{lem:valid-swap-exists} then guarantees the existence of an adjacent $\nearrow$ pair, so FBS can select one. By Observation~\ref{obs:single-swap}, swapping it decreases $N_{\nearrow}$ by~$1$ and increases $N_{\searrow}$ by~$1$, giving $N_{\nearrow}-N_{\searrow}=\Delta_0-2(k{+}1)$.

After $K^{\ast}$ swaps, $\Delta=\Delta_0-2K^{\ast}\in\{0,1\}\subseteq[0,\Delta^{\ast}]$ (using $\Delta^{\ast}\ge\Delta_0\bmod 2$), and the algorithm terminates. The greedy therefore uses exactly $K^{\ast}$ swaps, matching the lower bound.
\end{proof}

\subsection{Proof of Theorem~\ref{thm:unweighted-optimality} (Optimality for the unweighted case)}

\begin{proof}[Proof of Theorem~\ref{thm:unweighted-optimality}]
Unlike the weighted case, the unweighted Kemeny distance depends only on the \emph{number} of inversions, not on which specific pairs are inverted:
\[
d(r_s,r)=\frac{|\mathrm{Inv}(r_s,r)|}{\binom{n}{2}}.
\]
If $\Delta_0\le\Delta^{\ast}$, then $r_s$ is already feasible and $d(r_s,r_s)=0$ is trivially optimal. We therefore assume $\Delta_0>\Delta^{\ast}$, so that $K^{\ast}=\lceil(\Delta_0-\Delta^{\ast})/2\rceil\ge 1$. It suffices to show that $|\mathrm{Inv}(r_s,r_{\mathrm{FBS}})|=K^{\ast}$ and that $|\mathrm{Inv}(r_s,r)|\ge K^{\ast}$ for every feasible~$r$.

\emph{Lower bound.} Fix any feasible $r$, i.e.\ any $r$ with $\phi(r)\ge\psi$. By Lemma~\ref{lem:contiguous}, the minimum number of adjacent swaps taking $r_s$ to $r$ is $|\mathrm{Inv}(r_s,r)|$. This minimum-length sequence is an algorithm in $A_{swap}$ whose output is the feasible ranking~$r$, so by Proposition~\ref{prop:minimal-swaps} it uses at least $K^{\ast}$ swaps. Therefore
\[
|\mathrm{Inv}(r_s,r)|\ \ge\ K^{\ast}\qquad\text{for every feasible }r.
\]

\emph{FBS matches the bound.} By Proposition~\ref{prop:minimal-swaps}, FBS terminates after exactly $K^{\ast}$ adjacent swaps at a ranking $r_{\mathrm{FBS}}$ with $\Delta(r_{\mathrm{FBS}})\le\Delta^{\ast}$, so $r_{\mathrm{FBS}}$ is feasible. By Observation~\ref{obs:single-swap}, each adjacent swap creates or resolves exactly one $r_s$-inversion, so $|\mathrm{Inv}(r_s,\cdot)|$ changes by exactly $\pm 1$ per swap. Since $|\mathrm{Inv}(r_s,r_s)|=0$, we have
\[
|\mathrm{Inv}(r_s,r_{\mathrm{FBS}})|\ \le\ K^{\ast}.
\]
Combined with the lower bound (applied to the feasible $r_{\mathrm{FBS}}$) we get $|\mathrm{Inv}(r_s,r_{\mathrm{FBS}})|=K^{\ast}$, and hence
\[
d(r_s,r_{\mathrm{FBS}})=\frac{K^{\ast}}{\binom{n}{2}}=\min_{r:\phi(r)\ge\psi} d(r_s,r). \qedhere
\]
\end{proof}

\section{Proofs of Section 6}\label{app:proofs_weighted}

\subsection{Proof of Proposition~\ref{prop:weighted-approx} (Weight-spread approximation bound)}

\begin{proof}[Proof of Proposition~\ref{prop:weighted-approx}]
If $\Delta_0\le\Delta^{\ast}$, then $r_s$ is already feasible, $K^{\ast}=0$, and $d(r_s,r_{\mathrm{FBS}})=d(r_s,r^{\ast})=0$; the bound holds trivially. We therefore assume $\Delta_0>\Delta^{\ast}$, so that $K^{\ast}\ge 1$.

\emph{Upper bound on $d(r_s,r_{\mathrm{FBS}})$.} By Proposition~\ref{prop:minimal-swaps}, FBS performs exactly $K^{\ast}$ adjacent swaps starting from~$r_s$. By Observation~\ref{obs:single-swap}, each swap changes $|\mathrm{Inv}(r_s,\cdot)|$ by exactly~$\pm 1$. Since $|\mathrm{Inv}(r_s,r_s)|=0$ and $r_{\mathrm{FBS}}$ is feasible (hence $|\mathrm{Inv}(r_s,r_{\mathrm{FBS}})|\ge K^{\ast}$ by Lemma~\ref{lem:contiguous} and Proposition~\ref{prop:minimal-swaps}), we obtain $|\mathrm{Inv}(r_s,r_{\mathrm{FBS}})|=K^{\ast}$ (in particular, every swap creates a new $r_s$-inversion). Therefore
\[
d(r_s,r_{\mathrm{FBS}})=\sum_{(i,j)\in\mathrm{Inv}(r_s,r_{\mathrm{FBS}})}w_{ij}\ \le\ K^{\ast}\cdot w_{\max}.
\]

\emph{Lower bound on $d(r_s,r^{\ast})$.} By Lemma~\ref{lem:contiguous} and Proposition~\ref{prop:minimal-swaps}, every feasible ranking $r$ satisfies $|\mathrm{Inv}(r_s,r)|\ge K^{\ast}$. Hence
\[
d(r_s,r^{\ast})=\sum_{(i,j)\in\mathrm{Inv}(r_s,r^{\ast})}w_{ij}\ \ge\ K^{\ast}\cdot w_{\min}.
\]

Dividing the upper by the lower bound (both positive since $K^{\ast}\ge 1$ and $w_{\min}>0$) gives the stated ratio.
\end{proof}

\paragraph{Score-rank convention.}
Throughout this appendix we write ``$s(i)=i$'' as shorthand for equally-spaced scores: only the score gaps enter the weights, yielding pair weights $w_{uv}=|s(u)-s(v)|=|u-v|$ up to a global normalisation that cancels in all ratios, so the labeling $r_s=\mathrm{id}$ (item~$i$ at position~$i$) is without loss of generality. Similarly, ``$a=s$'' refers to the configuration $a(i)=i$, which under $r_s=\mathrm{id}$ gives $\tau(r_s,a)=+1$ — the maximum-imbalance case in which all $\binom{n}{2}$ pairs are concordant.

\subsection{Proof of Lemma~\ref{lem:spearman-kendall} (Spearman--Kendall identity)}\label{app:identity-proof}

\begin{proof}[Proof of Lemma~\ref{lem:spearman-kendall}]
Items are labelled $1,\ldots,n$ with scores $s(i)=i$ and unnormalised pair weights $w_{uv}=v-u$ for $u<v$ (the global factor $1/W$ cancels in all ratios and is omitted throughout this appendix). For a ranking~$r$, we must show
\begin{equation}\label{eq:app-identity}
\sum_{\substack{u<v\\r(u)>r(v)}}(v-u)\ =\ \sum_i i^2-\sum_i i\,r(i)\ =\ \tfrac{1}{2}\sum_i(r(i)-i)^2.
\end{equation}
The second equality follows from $\sum r(i)^2=\sum i^2$ (both range over $\{1,\ldots,n\}$). For the first equality, define for each item~$i$:
\begin{align*}
\ell(i)&:=\#\{j<i:r(j)>r(i)\}\quad\text{(left inversions)},\\
\rho(i)&:=\#\{j>i:r(j)<r(i)\}\quad\text{(right inversions)}.
\end{align*}

\paragraph{Step 1: $\ell(i)-\rho(i)=i-r(i)$.}
Among items $\{1,\ldots,i{-}1\}$, let $p$ denote how many occupy positions $<r(i)$. Then $\ell(i)=(i{-}1)-p$. Since exactly $r(i){-}1$ items in total sit before position~$r(i)$, $p+\rho(i)=r(i){-}1$, giving $p=r(i){-}1-\rho(i)$. Substituting: $\ell(i)=(i{-}1)-(r(i){-}1-\rho(i))=i-r(i)+\rho(i)$, hence $\ell(i)-\rho(i)=i-r(i)$.

\paragraph{Step 2: The right-hand side equals $\sum_i i\,(\ell(i)-\rho(i))$.}
Rewriting the inversion sum:
\[
\sum_{\substack{u<v\\r(u)>r(v)}}(v-u)=\sum_v v\,\ell(v)-\sum_u u\,\rho(u)=\sum_i i\,\bigl(\ell(i)-\rho(i)\bigr).
\]

\paragraph{Conclusion.}
By Step~1, $\sum_i i(\ell(i)-\rho(i))=\sum_i i(i-r(i))=\sum_i i^2-\sum_i i\,r(i)$.
\end{proof}

\paragraph{Remark.}
The identity connects two classical rank-correlation measures: the weighted Kendall distance (with weights $v{-}u$) equals half the Spearman squared displacement. This is specific to equally-spaced (integer) scores and does not hold for general score arrays.

\subsection{Proof of Proposition~\ref{prop:ascending-bound} (Tight bound for the all-ascending case)}\label{app:five-block}

The proof of Proposition~\ref{prop:ascending-bound} rests on two ingredients: (i)~an exact characterisation of the FBS output and its closed-form cost for the all-ascending case $a=s$ (Proposition~\ref{thm:five-block}), and (ii)~an LP relaxation lower bound on~$d^{\ast}$ (Proposition~\ref{prop:lp-structure}). We develop each in turn, then combine them to prove, for this special case, the asymptotic convergence of the ratio $d_{\mathrm{FBS}}/\mathrm{LP}$ to $(1{+}\sqrt{2})/2$ and the strict $2$-approximation $d_{\mathrm{FBS}}<2\,d^{\ast}$.

\paragraph{Setup.}
When $a=s=(1,2,\ldots,n)$, all $\binom{n}{2}$ pairs are concordant. With $\psi=1$, FBS performs $K^{\ast}=\lceil\binom{n}{2}/2\rceil$ swaps. We restrict to $n=2^t$ ($t\ge 2$) where $\binom{n}{2}$ is even and $K^{\ast}=n(n{-}1)/4$. Write $m:=n/2$.

\begin{proposition}[FBS output and closed-form cost for $a{=}s$, $n{=}2^t$]\label{thm:five-block}
When $a=s=(1,\ldots,n)$ and $n=2^t\ge 4$, FBS produces the ranking~$r_{\mathrm{FBS}}$ defined by the following five blocks:
\begin{enumerate}
\item[\emph{(B1)}] Positions $1,\ldots,m/2$: items $m,\,m{-}1,\,\ldots,\,m/2{+}1$ \emph{(reversed upper-left quarter)};
\item[\emph{(B2)}] Position $m/2{+}1$: item $n$;
\item[\emph{(B3)}] Positions $m/2{+}2,\ldots,m{+}1$: items $m/2,\,m/2{-}1,\,\ldots,\,1$ \emph{(reversed lower-left quarter)};
\item[\emph{(B4)}] Position $m{+}2$: item $n{-}1$;
\item[\emph{(B5)}] Positions $m{+}3,\ldots,2m$: items $n{-}2,\,n{-}3,\,\ldots,\,m{+}1$ \emph{(reversed right half)}.
\end{enumerate}
The total FBS cost is
\begin{equation}\label{eq:dgr-closed}
d_{\mathrm{FBS}}(n)\ =\ \frac{m(8m^2+21m-14)}{24},\qquad m=n/2.
\end{equation}
\end{proposition}

We prove the two parts of the proposition separately. We first derive the closed-form cost~\eqref{eq:dgr-closed} assuming the five-block structure, and then prove the five-block structure itself by induction on~$t$.

\begin{proof}[Proof of Eq.~\eqref{eq:dgr-closed}]
Assuming the five-block structure, we compute $d_{\mathrm{FBS}}=\tfrac{1}{2}S_2(r_{\mathrm{FBS}})$ using Lemma~\ref{lem:spearman-kendall}. The displacement of item~$i$ is $\delta_i:=r_{\mathrm{FBS}}(i)-i$. We evaluate $S_2=\sum_i\delta_i^2$ block by block.

\emph{Block~B1} (positions $j=1,\ldots,m/2$, item $m{+}1{-}j$).
Displacement $\delta_j=m{+}1{-}2j$, taking all odd values from $m{-}1$ down to~$1$. Hence $S_2^{(1)}=m(m^2{-}1)/6$.

\emph{Block~B2} (position $m/2{+}1$, item $2m$).
$S_2^{(2)}=(3m/2{-}1)^2$.

\emph{Block~B3} (positions $m/2{+}2,\ldots,m{+}1$). $\delta=2k$ for $k=1,\ldots,m/2$, giving $S_2^{(3)}=m(m{+}1)(m{+}2)/6$.

\emph{Block~B4} (position $m{+}2$, item $2m{-}1$). $S_2^{(4)}=(m{-}3)^2$.

\emph{Block~B5} (positions $m{+}3,\ldots,2m$, items $2m{-}2,\ldots,m{+}1$). Item $i=3m{+}1{-}j$ sits at position~$j$, giving $\delta=2j{-}3m{-}1$. As $j$ ranges from $m{+}3$ to $2m$, these are the $m{-}2$ consecutive odd values $5{-}m,\,7{-}m,\,\ldots,\,m{-}3,\,m{-}1$. The full set of odd values from $-(m{-}1)$ to $m{-}1$ has $m$~elements with sum of squares $m(m^2{-}1)/3$; Block~B5 omits the two largest absolute values $m{-}1$ and $m{-}3$ (which belong to the items in blocks B1 and B4 respectively). Hence $S_2^{(5)}=m(m^2{-}1)/3-(m{-}1)^2-(m{-}3)^2$.

\emph{Total.} The $(m{-}3)^2$ terms from B4 and B5 cancel. Combining the three fractional terms:
\begin{align*}
\frac{m(m^2{-}1)}{6}+\frac{m(m{+}1)(m{+}2)}{6}+\frac{m(m^2{-}1)}{3}
&=\frac{m(m{+}1)\bigl[(m{-}1)+(m{+}2)+2(m{-}1)\bigr]}{6}\\
&=\frac{m(m{+}1)(4m{-}1)}{6}.
\end{align*}
The remaining squared terms give $(3m/2{-}1)^2-(m{-}1)^2=m(5m{-}4)/4$. Therefore
\begin{align*}
S_2&=\frac{m(m{+}1)(4m{-}1)}{6}+\frac{m(5m{-}4)}{4}
=\frac{2m(m{+}1)(4m{-}1)+3m(5m{-}4)}{12}\\
&=\frac{m(8m^2+21m{-}14)}{12},
\end{align*}
where the last step uses $2(m{+}1)(4m{-}1)=8m^2{+}6m{-}2$ and $3(5m{-}4)=15m{-}12$, giving $8m^2{+}21m{-}14$. Hence $d_{\mathrm{FBS}}=S_2/2=m(8m^2+21m{-}14)/24$.
\end{proof}

\paragraph{Proof of the five-block structure.}
The proof proceeds by induction on~$t$, exploiting a hierarchical merge structure forced by FBS's min-cost rule. We establish the required auxiliary definitions and lemmas, then give the main induction.

\begin{definition}[Reversed block]\label{def:rev-block}
A \emph{reversed block of size~$b$} at positions $p,p{+}1,\ldots,p{+}b{-}1$ containing items $\{q,q{+}1,\ldots,q{+}b{-}1\}$ means that position $p{+}j$ holds item $q{+}b{-}1{-}j$ for $j=0,\ldots,b{-}1$.
\end{definition}

\begin{lemma}[Phase~1: disjoint transpositions]\label{lem:phase1}
The first $m$ FBS swaps transpose the pairs at positions $(1,2),(3,4),\ldots,(n{-}1,n)$, each at cost~$1$. After phase~$1$, the state consists of $m$ reversed blocks of size~$2$: block~$j$ at positions $2j{-}1,2j$ contains items $\{2j{-}1,2j\}$ in reversed order, for $j=1,\ldots,m$.
\end{lemma}

\begin{proof}
Initially all $n{-}1$ adjacent pairs are concordant with cost~$1$. FBS picks position~$1$ (leftmost minimum-cost eligible position). After swapping positions $1$ and $2$: position~$1$ becomes discordant ($\sigma[1]={-}1$); position~$2$ has cost $|s(1){-}s(3)|=2$; all positions $k\ge 3$ remain at cost~$1$. Hence the next FBS choice is position~$3$. By induction on $j$: after $j$ transpositions, the first $2j$ positions form reversed pairs with discordant internal signs, while positions $2j{+}1,\ldots,n{-}1$ remain concordant at cost~$1$. FBS next selects position $2j{+}1$. After all $m$ transpositions, every position $2j{-}1$ holds the pair $(2j,2j{-}1)$ and no cost-$1$ eligible positions remain.
\end{proof}

\begin{lemma}[Post-phase-$1$ eligible positions]\label{lem:post-phase1}
After phase~$1$, the eligible (concordant) adjacent positions are precisely the even indices $k=2,4,\ldots,n{-}2$, each with cost~$3$. There are $m{-}1$ such positions.
\end{lemma}

\begin{proof}
At even position $k=2j$, the items are $2j{-}1$ (right element of block~$j$) and $2j{+}2$ (left element of block~$j{+}1$). Since $2j{-}1<2j{+}2$, the pair is concordant with cost $|s(2j{-}1){-}s(2j{+}2)|=3$. At odd positions, items form a reversed pair within a block and are discordant.
\end{proof}

\begin{lemma}[Merge Lemma]\label{lem:merge}
Suppose the current state contains two adjacent reversed blocks of size~$b$: a \emph{left block} at positions $p,\ldots,p{+}b{-}1$ containing items $\{q,\ldots,q{+}b{-}1\}$ reversed, and a \emph{right block} at positions $p{+}b,\ldots,p{+}2b{-}1$ containing items $\{q{+}b,\ldots,q{+}2b{-}1\}$ reversed. Suppose furthermore that:
\begin{enumerate}
\item[\emph{(i)}] the boundary position $p{+}b{-}1$ (between items $q$ and $q{+}2b{-}1$) is the unique eligible position with minimum cost within these $2b$ positions;
\item[\emph{(ii)}] no eligible position outside these $2b$ positions has cost less than the boundary cost $2b{-}1$.
\end{enumerate}
Then FBS performs exactly $b^2$ swaps within these $2b$ positions, merging them into a single reversed block of size~$2b$ containing items $\{q,\ldots,q{+}2b{-}1\}$ in decreasing order. The total cost of these $b^2$ swaps is~$b^3$. The merge proceeds in $b$~rounds, where round~$r$ ($r=0,\ldots,b{-}1$) consists of $b$ swaps with costs $2b{-}1{-}r,\;2b{-}2{-}r,\;\ldots,\;b{-}r$.
\end{lemma}

\begin{proof}
We prove by induction on $r$ ($r={-}1,0,\ldots,b{-}1$, where $r={-}1$ denotes the initial state) the following invariant.

\smallskip\noindent\textbf{Invariant after round~$r$:} relative to local positions $0,\ldots,2b{-}1$ (offset by~$p$), the state consists of three segments:
\begin{itemize}
\item \textbf{Segment~1} (positions $0,\ldots,r$): items $q{+}2b{-}1,\,q{+}2b{-}2,\,\ldots,\,q{+}2b{-}1{-}r$ in decreasing order.
\item \textbf{Segment~2} (positions $r{+}1,\ldots,b{+}r$): items $q{+}b{-}1,\,q{+}b{-}2,\,\ldots,\,q$ in decreasing order (the original left-block items, shifted right by $r{+}1$).
\item \textbf{Segment~3} (positions $b{+}r{+}1,\ldots,2b{-}1$): items $q{+}2b{-}2{-}r,\,\ldots,\,q{+}b$ in decreasing order (the remaining right-block items).
\end{itemize}
All adjacent pairs within each segment are discordant. The unique eligible position with minimum cost (within these $2b$ positions) is the boundary between segments~2 and~3 at position $b{+}r$, with cost $2b{-}2{-}r$.

\smallskip\noindent\textbf{Base case ($r={-}1$):} Segment~1 is empty, segment~2 is the left block, segment~3 is the right block. The only eligible position is the boundary at position~$b{-}1$, with cost $2b{-}1$.

\smallskip\noindent\textbf{Round $r{+}1$ execution:} The boundary between segments~2 and~3 is at position $b{+}r$, where item~$q$ sits at position $b{+}r$ and item $q{+}2b{-}2{-}r$ sits at position $b{+}r{+}1$. FBS swaps these two items at cost $2b{-}2{-}r$.

After this initial swap, item $q{+}2b{-}2{-}r$ at position $b{+}r$ propagates leftward through segment~2. At each step the leftward swap is the cheapest eligible move: the propagation front has strictly decreasing cost ($2b{-}3{-}r,\,2b{-}4{-}r,\,\ldots$), while the only other eligible position---item~$q$ adjacent to segment~3's leftmost item at position $b{+}r{+}1$---has fixed cost $2b{-}3{-}r$. At the first step both costs are equal and leftmost tie-breaking selects the propagation; at every subsequent step the leftward cost is strictly smaller.

After $b{-}1$ leftward swaps, item $q{+}2b{-}2{-}r$ reaches position $r{+}1$, adjacent to item $q{+}2b{-}1{-}r$ in segment~1; since $q{+}2b{-}2{-}r<q{+}2b{-}1{-}r$, this pair is discordant. The state matches the invariant with $r$ replaced by $r{+}1$. Total swaps in round $r{+}1$: $1+(b{-}1)=b$, with costs $2b{-}1{-}(r{+}1),\;2b{-}2{-}(r{+}1),\;\ldots,\;b{-}(r{+}1)$.

\smallskip\noindent\textbf{After round $b{-}1$:} Segment~1 spans all $2b$ positions with items $q{+}2b{-}1,\ldots,q$ in decreasing order; segments~2 and~3 are empty. The merge is complete.

\smallskip\noindent\textbf{Cost calculation:}
Each round~$r$ sums an arithmetic series of $b$~terms from $b{-}r$ to $2b{-}1{-}r$, giving per-round total $b(3b{-}1{-}2r)/2$. Summing over $r=0,\ldots,b{-}1$ yields another arithmetic series from $b(3b{-}1)/2$ to $b(b{+}1)/2$, so the total cost is $b\cdot\frac{b(3b{-}1)/2+b(b{+}1)/2}{2}=b^3$.
The swap count is $b\cdot b=b^2$.
\end{proof}

\begin{lemma}[Level separation]\label{lem:level-sep}
At the start of level~$\ell$ ($\ell=1,\ldots,t{-}2$), the state consists of $n/b=2^{t-\ell}$ reversed blocks of size $b=2^\ell$, where block~$j$ contains items $\{(j{-}1)b{+}1,\ldots,jb\}$. All boundary costs between adjacent blocks equal $2b{-}1$. Consequently, FBS completes all merges at level~$\ell$ before touching any boundary at a higher level, and merges proceed left to right within each level.
\end{lemma}

\begin{proof}
At the start of level~$\ell$, the boundary between any two adjacent blocks has cost $2b{-}1$: the right end of each block holds its smallest item and the left end holds its largest, separated by $2b{-}1$.

After the leftmost merge (blocks $1,2$) completes, the merged block of size~$2b$ is fully reversed: all internal pairs are discordant and ineligible. The boundary between the merged block and the next unmerged block~$3$ has cost $3b{-}1$: the merged block's smallest item is~$1$ (at its right end) and block~$3$'s largest item is~$3b$ (at its left end), giving $3b-1$. Since $3b{-}1>2b{-}1$ (for $b\ge 1$), FBS proceeds to the next intra-pair boundary (blocks $3,4$, still at cost $2b{-}1$). The argument generalises: after any merge pair completes, the inter-pair boundary has cost $3b{-}1>2b{-}1$, so merges proceed left to right.

After all merges at level~$\ell$ complete, the state consists of $n/(2b)$ reversed blocks of size~$2b$, each with boundary cost $4b{-}1=2(2b){-}1$---precisely the starting configuration for level~$\ell{+}1$.
\end{proof}

\begin{proof}[Proof of Proposition~\ref{thm:five-block} (five-block structure)]
We proceed by induction on $t$.

\smallskip\noindent\textbf{Base case ($t=2$, $n=4$).}
$K^{\ast}=3$. The FBS trace is: swap positions $(1,2)$ at cost~$1$, then $(3,4)$ at cost~$1$, then $(2,3)$ at cost~$3$, producing $(2,4,1,3)$. This matches: B1$=\{2\}$, B2$=\{4\}$, B3$=\{1\}$, B4$=\{3\}$, B5 empty.

\smallskip\noindent\textbf{Inductive step ($t\ge 3$).}

\emph{Phase~1.} By Lemma~\ref{lem:phase1}, the first $m$ swaps produce $m$ reversed blocks of size~$2$.

\emph{Levels $1$ through $t{-}2$ (full merges).} By Lemma~\ref{lem:post-phase1}, after phase~1 the eligible positions are the $m{-}1$ even indices, all at cost $3=2\cdot 2{-}1$. These are the intra-pair boundaries for level-$1$ merges. By Lemma~\ref{lem:level-sep}, FBS processes merges left to right at each level. By the Merge Lemma~\ref{lem:merge}, each level-$\ell$ merge of two blocks of size $b=2^\ell$ uses $b^2$ swaps at cost~$b^3$. At level~$\ell$ there are $n/(2b)$ merges, giving $2^{t-\ell-1}\cdot b^2=2^{t+\ell-1}$ swaps per level.

After level~$t{-}2$, the state consists of two reversed blocks of size~$m$: the left block at positions $1,\ldots,m$ with items $m,\ldots,1$, and the right block at positions $m{+}1,\ldots,2m$ with items $2m,\ldots,m{+}1$.

\emph{Swap budget.} Total swaps through level $t{-}2$: $\sum_{k=t-1}^{2t-3}2^k=2^{t-1}(2^{t-1}{-}1)=m(m{-}1)$. Since $K^{\ast}=m(2m{-}1)/2$, exactly $K^{\ast}-m(m{-}1)=m/2$ swaps remain.

\emph{Top level (partial merge).} The boundary at position~$m$ has cost $2m{-}1$. By the Merge Lemma's round-$0$ structure, item~$n=2m$ (leftmost element of the right block) propagates leftward from position $m{+}1$ through positions $m,\,m{-}1,\,\ldots,\,m/2{+}1$, displacing items $1,2,\ldots,m/2$ one position to the right. This uses exactly $m/2$ swaps (the remaining budget) at costs $2m{-}1,\,2m{-}2,\,\ldots,\,3m/2$. The algorithm terminates because after $K^{\ast}$ total swaps, $\Delta=\binom{n}{2}-2K^{\ast}=0\le\Delta^{\ast}$.

\emph{Final state.} After the partial merge:
\begin{itemize}
\item Positions $1,\ldots,m/2$: items $m,m{-}1,\ldots,m/2{+}1$ (undisturbed upper portion of the left block). This is \textbf{B1}.
\item Position $m/2{+}1$: item $2m$ (propagated here). This is \textbf{B2}.
\item Positions $m/2{+}2,\ldots,m{+}1$: items $m/2,m/2{-}1,\ldots,1$ (lower portion of the left block, shifted right by one). This is \textbf{B3}.
\item Position $m{+}2$: item $2m{-}1$ (second element of the right block, unmoved since only item~$2m$ propagated left). This is \textbf{B4}.
\item Positions $m{+}3,\ldots,2m$: items $2m{-}2,\ldots,m{+}1$ (remainder of the right block). This is \textbf{B5}.
\end{itemize}
This matches the five-block structure of Proposition~\ref{thm:five-block}.
\end{proof}

\paragraph{LP relaxation lower bound.}\label{app:lp-relaxation}
To compare the FBS cost against a lower bound, we relax the constraint that the inverted pairs form a valid inversion set (i.e., arise from a permutation). Let $P^{\nearrow}:=\{(i,j):i<j,\,\alpha_{ij}=+1\}$ denote the set of concordant pairs in $r_s$. Introducing indicators $x_{ij}\in[0,1]$ for each concordant pair and dropping the transitivity constraint gives the \emph{fractional knapsack} relaxation:
\[
\mathrm{LP}:=\min\sum_{(i,j)\in P^{\nearrow}}w_{ij}\,x_{ij}\qquad\text{s.t.}\quad\sum_{(i,j)\in P^{\nearrow}}x_{ij}=K^{\ast},\quad 0\le x_{ij}\le 1.
\]
The solution is immediate: set $x_{ij}=1$ for the $K^{\ast}$ pairs with smallest weight. Hence $\mathrm{LP}=\sum_{k=1}^{K^{\ast}}w_{(k)}$, where $w_{(1)}\le w_{(2)}\le\cdots$ are the sorted pair weights. Since $\mathrm{LP}$ relaxes constraints present in the true optimisation, $\mathrm{LP}\le d^{\ast}$.

\begin{proposition}[LP closed form for $a=s$]\label{prop:lp-structure}
For integer scores $s(i)=i$ with all pairs concordant, there are exactly $(n{-}d)$ pairs of weight~$d$ for each $d=1,\ldots,n{-}1$. The LP selects pairs greedily from cheapest to most expensive:
\[
\mathrm{LP}(n) = \sum_{d=1}^{D-1}d\,(n{-}d) + D\cdot R,
\]
where $D$ is the smallest integer with $\sum_{d=1}^{D}(n{-}d)\ge K^{\ast}$ and $R:=K^{\ast}-\sum_{d=1}^{D-1}(n{-}d)$.
\end{proposition}

\begin{proof}[Proof of Proposition~\ref{prop:lp-structure}]
For $s(i)=i$, each pair $(i,j)$ with $i<j$ has weight $w_{ij}=j{-}i=:d$. For each $d\in\{1,\ldots,n{-}1\}$ there are exactly $n{-}d$ pairs at distance~$d$ (namely $(i,i{+}d)$ for $i=1,\ldots,n{-}d$). The LP selects the $K^{\ast}$ cheapest pairs, so it fills all $n{-}1$ pairs of weight~$1$, then all $n{-}2$ pairs of weight~$2$, and so on through weight~$D{-}1$, consuming $\sum_{d=1}^{D-1}(n{-}d)$ pairs at total cost $\sum_{d=1}^{D-1}d(n{-}d)$. The remaining $R=K^{\ast}-\sum_{d=1}^{D-1}(n{-}d)$ pairs are taken from the $n{-}D$ available pairs of weight~$D$ (note $R\le n{-}D$ by minimality of~$D$), contributing $D\cdot R$.
\end{proof}

The critical weight satisfies $D=(2{-}\sqrt{2})\,m+O(1)$ as $m\to\infty$: solving $\sum_{d=1}^{D}(n{-}d)=K^{\ast}$, i.e.\ $Dn-D(D{+}1)/2=m(2m{-}1)/2$, gives $D^2-2(2m{-}1)D+m(2m{-}1)=0$, whose positive root is $(2{-}\sqrt{2})m+O(1)$. Despite being called ``LP'', the value requires no solver and is computed in $O(n)$ time.

\begin{proof}[Proof of Proposition~\ref{prop:ascending-bound} (Tight bound for the all-ascending case)]
Write $m=n/2$ and $c=2-\sqrt{2}$.

\emph{FBS cost.} By Proposition~\ref{thm:five-block}, $d_{\mathrm{FBS}}=m(8m^2+21m-14)/24=m^3/3+O(m^2)$.

\emph{LP lower bound.} By Proposition~\ref{prop:lp-structure}, $\mathrm{LP}=\sum_{d=1}^{D-1}d(n{-}d)+D\cdot R$ with $D=cm+O(1)$. Expanding: $\sum_{d=1}^{D-1}d(2m{-}d)=c^2(1{-}c/3)\,m^3+O(m^2)$. The remainder term $D\cdot R$ contributes $O(m^2)$, so $\mathrm{LP}=c^2(1{-}c/3)\,m^3+O(m^2)$.

\emph{Convergence.} The ratio satisfies
\[
\frac{d_{\mathrm{FBS}}}{\mathrm{LP}}=\frac{m^3/3+O(m^2)}{c^2(1{-}c/3)\,m^3+O(m^2)}\;\longrightarrow\;\frac{1}{3c^2(1{-}c/3)}.
\]
Substituting $c=2{-}\sqrt{2}$: $c^2=6{-}4\sqrt{2}$, $1{-}c/3=(1{+}\sqrt{2})/3$, so $3c^2(1{-}c/3)=(6{-}4\sqrt{2})(1{+}\sqrt{2})=6{+}6\sqrt{2}{-}4\sqrt{2}{-}8=2(\sqrt{2}{-}1)$. Hence the limit is $1/[2(\sqrt{2}{-}1)]=(1{+}\sqrt{2})/2\approx 1.207$.

\emph{Strict 2-approximation.} Since $\mathrm{LP}\le d^{\ast}$, it suffices to show $d_{\mathrm{FBS}}<2\,\mathrm{LP}$ for all $t\ge 2$. For even $m\ge 2$, the $m/2$ cheapest weight classes contain $\sum_{d=1}^{m/2}(2m{-}d)=m(7m{-}2)/8\le m(2m{-}1)/2=K^{\ast}$ pairs (the inequality reduces to $m\ge 2$), so
\[
\mathrm{LP}\;\ge\;\sum_{d=1}^{m/2}d\,(2m{-}d)\;=\;\frac{m(m{+}2)(5m{-}1)}{24}.
\]
The condition $d_{\mathrm{FBS}}<2\,\mathrm{LP}$ then becomes
\[
\frac{m(8m^2{+}21m{-}14)}{24}\;<\;\frac{m(m{+}2)(5m{-}1)}{12},
\]

which simplifies to $2m^2{-}3m{+}10>0$. Since the discriminant $9{-}80=-71<0$ and the leading coefficient is positive, this holds for all real~$m$.
\end{proof}

\begin{table}[h]
\centering
\begin{tabular}{@{}rrrrr@{}}
\toprule
$t$ & $n=2^t$ & $d_{\mathrm{FBS}}$ & $\mathrm{LP}$ & $d_{\mathrm{FBS}}/\mathrm{LP}$\\
\midrule
2 & 4 & 5 & 3 & 1.667\\
3 & 8 & 33 & 22 & 1.500\\
4 & 16 & 222 & 160 & 1.387\\
5 & 32 & 1\,580 & 1\,205 & 1.311\\
6 & 64 & 11\,800 & 9\,348 & 1.262\\
7 & 128 & 90\,928 & 73\,587 & 1.236\\
8 & 256 & 713\,312 & 583\,900 & 1.222\\
9 & 512 & 5\,649\,600 & 4\,652\,075 & 1.214\\
10 & 1\,024 & 44\,968\,320 & 37\,139\,950 & 1.211\\
11 & 2\,048 & 358\,830\,848 & 296\,812\,700 & 1.209\\
12 & 4\,096 & 2\,866\,980\,352 & 2\,373\,273\,400 & 1.208\\
13 & 8\,192 & 22\,921\,169\,920 & 18\,981\,273\,200 & 1.208\\
15 & 32\,768 & $1.47\times 10^{12}$ & $1.21\times 10^{12}$ & 1.207\\
17 & 131\,072 & $9.38\times 10^{13}$ & $7.77\times 10^{13}$ & 1.207\\
19 & 524\,288 & $6.00\times 10^{15}$ & $4.97\times 10^{15}$ & 1.207\\
21 & 2\,097\,152 & $3.84\times 10^{17}$ & $3.18\times 10^{17}$ & 1.207\\
\bottomrule
\end{tabular}
\caption{FBS cost vs.\ LP relaxation for $a=s$, $n=2^t$, $t=2,\ldots,21$. The ratio converges to $(1{+}\sqrt{2})/2\approx 1.207$.}
\label{tab:ratio-appendix}
\end{table}

\subsection{Towards a proof of the 2-approximation}\label{app:proof-strategy}

This appendix explains why the constant~$2$ in the 2-approximation conjecture is structurally motivated, outlines the proof path it suggests, and identifies the remaining open step.

\paragraph{Step 1: FBS and optimum live in the same Kendall ball.}
By Theorem~\ref{thm:unweighted-optimality}, every feasible ranking has at least $K^{\ast}$ inversions, and FBS achieves exactly $K^{\ast}$. Hence
\[
K(r_{\mathrm{FBS}})\;=\;K(r^{\ast})\;=\;K^{\ast},
\]
i.e.\ both rankings sit on the sphere $\{r:|\mathrm{Inv}(r_s,r)|=K^{\ast}\}$. The two solutions therefore differ only in \emph{which} $K^{\ast}$ pairs are inverted, not in their cardinality.

\paragraph{Step 2: Diaconis--Graham yields a factor-$2$ ratio on the Spearman footrule.}
The classical Diaconis--Graham inequality~\cite{diaconis1977spearman} states that for every ranking $r$,
\[
K(r)\;\le\;F(r)\;\le\;2K(r),\qquad F(r):=\sum_{i}|r(i)-i|=\|\boldsymbol{\delta}\|_1.
\]
Applying this to both rankings (which have $K=K^{\ast}$ by Step~1):
\[
\|\boldsymbol{\delta}_{\mathrm{FBS}}\|_1\;=\;F(r_{\mathrm{FBS}})\;\le\;2K^{\ast},\qquad \|\boldsymbol{\delta}^{\ast}\|_1\;=\;F(r^{\ast})\;\ge\;K^{\ast}.
\]
Dividing gives the \emph{footrule-ratio bound}
\[
\frac{\|\boldsymbol{\delta}_{\mathrm{FBS}}\|_1}{\|\boldsymbol{\delta}^{\ast}\|_1}\;\le\;2.
\]

\paragraph{Step 3: The Spearman--Kendall identity propagates the ratio to $d$.}
By Lemma~\ref{lem:spearman-kendall}, $d(r_s,r)=\tfrac{1}{2}\|\boldsymbol{\delta}\|_2^2$, so
\[
\frac{d(r_s,r_{\mathrm{FBS}})}{d(r_s,r^{\ast})}\;=\;\frac{\|\boldsymbol{\delta}_{\mathrm{FBS}}\|_2^2}{\|\boldsymbol{\delta}^{\ast}\|_2^2}.
\]
Converting the footrule ratio of Step~2 into a squared-displacement ratio is straightforward when the two displacement profiles are comparably spread. A standard tool is H\"older's inequality
\[
\|\boldsymbol{\delta}\|_2^2\;\le\;\|\boldsymbol{\delta}\|_\infty\cdot\|\boldsymbol{\delta}\|_1,
\]
which combined with Step~2 gives
\[
d(r_s,r_{\mathrm{FBS}})\;\le\;\tfrac{1}{2}\|\boldsymbol{\delta}_{\mathrm{FBS}}\|_\infty\cdot\|\boldsymbol{\delta}_{\mathrm{FBS}}\|_1\;\le\;\|\boldsymbol{\delta}_{\mathrm{FBS}}\|_\infty\cdot K^{\ast}.
\]
A factor-$2$ bound on $d$ then follows whenever $\|\boldsymbol{\delta}_{\mathrm{FBS}}\|_\infty/\|\boldsymbol{\delta}^{\ast}\|_\infty$ is bounded, i.e.\ whenever FBS does not concentrate its displacement on a single item disproportionately more than the optimum does.

\paragraph{Why the constant cannot be lowered through this route.}
The right-hand side $F\le 2K$ in the Diaconis--Graham inequality is \emph{tight}: it is saturated, for example, by the long shift $r=(2,3,\ldots,n,1)$, which has $K=n-1$ inversions and $F=2(n-1)$. Consequently, no proof going through Diaconis--Graham can yield a constant below~$2$. Tightening the conjecture would require an inequality that exploits the specific structure of $r_{\mathrm{FBS}}$---in particular, the fact that FBS's min-cost selection rule produces \emph{spread-out} displacement profiles, while the optimum may concentrate displacement on a few items.

\paragraph{The remaining open step: anti-concentration of FBS.}
The structural property still to be formalised is that FBS's min-cost rule prevents displacement concentration: an item that has already been displaced far creates expensive adjacent pairs, which FBS avoids in favour of cheaper swaps elsewhere. Concretely, we conjecture an \emph{injective charging argument}: every swap that contributes a large term to $\|\boldsymbol{\delta}_{\mathrm{FBS}}\|_2^2$ can be paired with at least one earlier swap of comparable cost in $r^{\ast}$. Combined with Steps~1--3 this would close the proof.

\paragraph{Empirical support.}
For the all-ascending case $a=s$ with $\psi=1$, the strict $2$-approximation is proved analytically (Proposition~\ref{prop:ascending-bound}), with the ratio converging to $(1{+}\sqrt{2})/2\approx 1.207$. For arbitrary attribute arrays, the exhaustive verification at $n\le 9$ (worst ratio~$1.46$) together with MILP-exact computation up to $n=100$ ($d_{\mathrm{FBS}}/d^{\ast}\le 1.31$ across $542$ instances; LP-bound $\le 1.32$ across $612$ instances, Appendix~\ref{app:milp}) strongly supports the conjecture, and indeed suggests that the sharp constant may be substantially below~$2$.

\section{Proof of Section 8}\label{app:proof-complexity}

\begin{proof}[Proof of Proposition~\ref{prop:complexity} (unweighted case)]
\textbf{Initialisation (lines~1--6).}
Sorting by score costs $O(n\log n)$.
Computing the sign array $\sigma$ costs $O(n)$.
The Kendall counts $N_{\nearrow}$ and $N_{\searrow}$ are each computed by merge-sort--based counting in $O(n\log n)$.
Building the linked list $L$ is a single pass over $\sigma$: $O(n)$.
Total initialisation: $O(n\log n)$.

\smallskip
\textbf{Main loop (lines~7--24).}
The loop executes exactly $K^{\ast}$ iterations (Proposition~\ref{prop:minimal-swaps}).
In each iteration: retrieving an element from $L$ is $O(1)$; the swap itself is $O(1)$; updating $\sigma$ for three neighbours and maintaining $L$ (at most $3$ insertions/deletions in a doubly-linked list) is $O(1)$; updating $N_{\nearrow},N_{\searrow}$ is $O(1)$.
Total per iteration: $O(1)$.
Total for the loop: $O(K^{\ast})$.

\smallskip
\textbf{Overall time (part~(i)).} $O(n\log n)+O(K^{\ast})=O(n\log n+K^{\ast})$.
In the worst case $K^{\ast}=\Theta(n^{2})$ and the $n\log n$ term is absorbed, giving $O(n^{2})$.

\smallskip
\textbf{Space (part~(ii)).}
The algorithm stores: the arrays $R[\cdot],A[\cdot]$ ($O(n)$); the sign array $\sigma[1..n{-}1]$ ($O(n)$); the doubly-linked list $L$ with at most $n{-}1$ entries ($O(n)$); and $O(1)$ scalars. Total: $O(n)$.
\end{proof}

\begin{proof}[Proof of Proposition~\ref{prop:complexity} (weighted case)]
\textbf{Initialisation (lines~1--7).}
\begin{itemize}
\item \emph{Line~1 (sort).} Sorting the $n$~columns by score costs $O(n\log n)$.
\item \emph{Lines~2--3 (adjacent costs and signs).} Each array has $n-1$ entries computed in $O(1)$ each, giving $O(n)$ total.
\item \emph{Line~4 (Kendall counts).} Computing $N_{\nearrow}=|\{i<j:A[i]<A[j]\}|$ and $N_{\searrow}=|\{i<j:A[i]>A[j]\}|$ independently, each by a merge-sort--based count, costs $O(n\log n)$.
\item \emph{Lines~5--6 (threshold and majority sign).} $O(1)$ each.
\item \emph{Line~7 (heap construction).} Build a min-heap $H$ over the $\le n{-}1$ entries $k$ with $\sigma[k]=\mu$, keyed by $(c[k],k)$. Bottom-up heap construction costs $O(n)$.
\end{itemize}
Total initialisation: $O(n\log n)$.

\smallskip
\textbf{Main loop (lines~8--28).}
By Proposition~\ref{prop:minimal-swaps} the loop executes exactly $K^{\ast}$ iterations. In each iteration:
\begin{itemize}
\item \emph{Line~9 (extract minimum).} $\textsc{Extract-Min}(H)$ returns $k^{\ast}$ in $O(\log n)$.
\item \emph{Line~10 (swap).} The swap itself is $O(1)$.
\item \emph{Lines~11--26 (neighbour updates and heap maintenance).} Exactly $3$ entries $\ell\in\{k^{\ast}{-}1,\,k^{\ast},\,k^{\ast}{+}1\}$ are inspected (fewer at the boundaries). For each:
    \begin{itemize}
    \item Recompute $c[\ell]$ and $\sigma[\ell]$ in $O(1)$.
    \item Perform at most one heap operation (insert, delete, or decrease-key), each costing $O(\log n)$.
    \end{itemize}
    Total per iteration: at most $3$ heap operations, costing $O(\log n)$.
\item \emph{Line~27 (Kendall counts).} The global counters $N_{\nearrow},N_{\searrow}$ are updated in $O(1)$.
\end{itemize}
Total for the loop: $O(\log n)\times K^{\ast}=O(K^{\ast}\log n)$.

\smallskip
\textbf{Overall time (part~(i)).} $O(n\log n)+O(K^{\ast}\log n)=O\!\bigl((n+K^{\ast})\log n\bigr)$.

\smallskip
\textbf{Worst case.} The maximum value of $K^{\ast}$ occurs when $\Delta_0=\binom{n}{2}$ (all pairs concordant, e.g.\ $a\equiv r_s$) and $\Delta^{\ast}=0$ (or~$1$), giving $K^{\ast}=\lfloor\binom{n}{2}/2\rfloor=\Theta(n^{2})$. The overall time is then $O(n^{2}\log n)$.

\smallskip
\textbf{Space (part~(ii)).}
The algorithm stores: the arrays $R[\cdot]$, $S[\cdot]$, $A[\cdot]$ (input, $O(n)$); the auxiliary arrays $c[1..n{-}1]$ and $\sigma[1..n{-}1]$ ($O(n)$ each); the min-heap $H$ with at most $n{-}1$ entries ($O(n)$); the scalars $N_{\nearrow}$, $N_{\searrow}$, $\Delta^{\ast}$, $k^{\ast}$, $\mu$ ($O(1)$). Total auxiliary space: $O(n)$.
\end{proof}

\section{Datasets}\label{app:datasets}

For the real-world experiments we use four datasets commonly adopted in the fair-ranking literature. For each, we report the chosen score, the chosen continuous protected attribute, the size~$n$ after removing missing entries, and the initial fairness $\phi_0$ (the value of~$\phi$ on the score-sorted ranking). All datasets are publicly available; preprocessing is intentionally minimal to keep results reproducible.

\paragraph{COMPAS~\citep{angwin2016machine}.}
The ProPublica COMPAS dataset of recidivism risk assessments. \emph{Score:} \texttt{decile\_score} (integer 1--10, higher = higher predicted risk). \emph{Continuous attribute:} \texttt{age} (years). After dropping rows with missing values, $n=7{,}214$ and $\phi_0=0.72$.

\paragraph{German Credit~\citep{statlog}.}
The Statlog (German Credit) dataset from the UCI repository. \emph{Score:} \texttt{credit\_amount} (Deutschmark, continuous). \emph{Continuous attribute:} \texttt{age} (years). $n=1{,}000$ (no missing values), $\phi_0=0.98$.

\paragraph{Law School~\citep{wightman1998lsac}.}
The LSAC National Longitudinal Bar Passage Study. \emph{Score:} \texttt{lsat} (LSAT score, continuous). \emph{Continuous attribute:} \texttt{ugpa} (undergraduate GPA, continuous). After dropping rows with missing values, $n=18{,}692$ and $\phi_0=0.85$.

\paragraph{Communities and Crime~\citep{communities_and_crime_183}.}
A community-level dataset from the UCI repository combining socio-economic data from the 1990 US Census, 1995 FBI Uniform Crime Report, and 1990 Law Enforcement Management and Administrative Statistics survey. All variables are normalized to $[0,1]$. \emph{Score:} \texttt{ViolentCrimesPerPop} (continuous, normalized violent-crime rate per population). \emph{Continuous attribute:} \texttt{PctPopUnderPov} (continuous, fraction of the population below the poverty line). $n=1{,}994$, $\phi_0=0.56$.

\paragraph{Score and attribute choice.}
For COMPAS and German Credit we follow the standard fair-ranking convention of using the dataset's primary risk/utility score and treating age as the continuous protected attribute. For Law School we follow the same convention used for predicting bar passage, with LSAT as the score and undergraduate GPA as the continuous attribute on which the ranking should remain balanced. For Communities and Crime, all numerical variables are continuous and pre-normalized; we choose the violent-crime rate as the score (the natural ranking target in this dataset) and the poverty rate as the protected attribute, motivated by the well-documented correlation between poverty and reported crime rates~\citep{kelly2000inequality}. This choice yields the lowest initial fairness among the four datasets ($\phi_0=0.56$) and therefore exhibits the richest fairness--cost trade-off curve.

\section{Baselines}\label{app:baselines}

We compare FBS against four baselines. Three are existing group-fair re-ranking algorithms that require a discrete protected group; we adapt them to the continuous setting by quantile-binning the attribute~$a$ into $G$ equal-frequency bins. The fourth, \emph{Blend}, is a continuous heuristic of our own that interpolates between the score order and a perfectly-fair target order. All baselines are evaluated on the same axes as FBS: achieved continuous fairness $\phi$ (computed on the original undiscretized attribute) and weighted Kemeny distance $d(r_s,r)$ from the score-based ranking.

\paragraph{DetConstSort~\citep{geyik2019fairness}.}
A deterministic constrained-sorting algorithm that enforces proportional representation of each protected group within every prefix of the output ranking. Given group proportions $p_g=|\{i:g(i)=g\}|/n$, it iteratively appends the highest-scoring remaining item subject to the constraint that, for every group~$g$, the count in the current prefix is at least $\lfloor p_g\cdot k\rfloor$ for prefix length~$k$. We use the implementation provided by FairRankTune~\citep{cachel2024fairranktune}. We report results for $G\in\{2,3,5,10\}$.

\paragraph{FA*IR~\citep{Zehlike2017}.}
A re-ranking algorithm that guarantees the protected group meets a minimum representation in every prefix, certified by a statistical test with significance~$\alpha=0.1$. FA*IR is intrinsically binary: a single ``protected'' group versus the rest. We restrict it to $G=2$ (median split on~$a$), assigning as protected the group most under-represented in the top half of the score-sorted ranking. Since the mtable computation is exponential in~$k$, in the Pareto experiments, we use $k=400$, as suggested by the authors. Furthermore, we use the implementation provided by fairsearchcore~\cite{fairsearchcore}.

\paragraph{Median-split interleaving (ours).}
A simple binary baseline that splits items at the median attribute value into two groups, sorts each group by descending score, and builds the output ranking by alternately picking the next highest-scored item from each group. When one group is exhausted, the remainder of the other is appended. This serves as a sanity-check baseline: it always achieves close-to-perfect attribute balance on the binary partition but ignores both score gaps and the continuous nature of~$a$.

\paragraph{Blend (continuous heuristic, ours).}
Unlike the above, Blend does not discretize~$a$. It computes, for each item~$i$:
(a)~its \emph{score-position} $\rho_s(i)$, the rank of~$i$ in the score-sorted order;
(b)~a \emph{fair-position} $\rho_f(i)$, obtained by sorting items by~$a$ and interleaving the high-attribute and low-attribute halves to produce a permutation with $\phi\approx 1$.
The output ranking is then obtained by sorting items by the convex combination
\[
\rho_\beta(i)\;=\;\beta\,\rho_s(i)\;+\;(1-\beta)\,\rho_f(i),\qquad \beta\in[0,1].
\]
At $\beta=1$ this reduces to the original score order ($\phi=\phi_0$, $d=0$); at $\beta=0$ it returns the perfectly-fair target order ($\phi\approx 1$, $d$ potentially very large). Sweeping~$\beta$ traces a curve in the fairness--cost plane. Because the blend is between integer ranks rather than scores, Blend is \emph{cost-blind}: a swap between two items with very different scores costs the same in the blend space as a swap between two near-equal-score items. This is the main reason Blend underperforms FBS on real data with skewed score distributions, where greedy minimum-cost swapping (FBS) keeps the cumulative cost much smaller for the same fairness gain.

\paragraph{Discretization information loss.}
The first three baselines depend on a discretization parameter~$G$. Increasing~$G$ allows finer fairness control but also increases the number of constraints and hence the cost of satisfying them. Crucially, even at $G=10$ the algorithm is blind to within-bin variation of~$a$, while FBS optimises the continuous fairness measure~$\phi$ at the pairwise level.

\section{MILP formulation for the exact optimum}\label{app:milp}

Section~\ref{sec:weighted-approx} (\emph{Exact MILP optimum for medium $n$}) reports the empirical approximation ratio $d_{\mathrm{FBS}}/d^{\ast}$ obtained by solving the weighted Kemeny problem with fairness constraint exactly. Here we describe the MILP formulation used.

For each unordered pair $(i,j)$ with $i<j$ we introduce a binary variable $x_{ij}\in\{0,1\}$ that equals $1$ if and only if the pair is inverted in the output ranking with respect to the score-sorted ranking $r_s$. Writing $\alpha_{ij}=\operatorname{sgn}(a(j)-a(i))\in\{-1,+1\}$ and $D_0=\sum_{i<j}\alpha_{ij}$, the problem reads:
\[
\min_{x\in\{0,1\}^{\binom{n}{2}}}\ \sum_{i<j}|s(i)-s(j)|\,x_{ij}\qquad\text{s.t.}\quad
\big|D_0-2\sum_{i<j}\alpha_{ij}\,x_{ij}\big|\ \le\ \Delta^{\ast},
\]
together with the standard FAS-style triangle (transitivity) constraints
\[
x_{ij}+x_{jk}-x_{ik}\le 1,\qquad -x_{ij}-x_{jk}+x_{ik}\le 0\qquad\forall\,i<j<k,
\]
which together rule out the only two binary inversion patterns inconsistent with a permutation. The fairness constraint is encoded as the pair of linear inequalities $\alpha^{\!\top} x\le(D_0+\Delta^{\ast})/2$ and $\alpha^{\!\top} x\ge(D_0-\Delta^{\ast})/2$.

For $n\in\{20,50,100\}$ this yields a MILP with $\binom{n}{2}$ binary variables and $2+2\binom{n}{3}$ linear constraints (e.g., $190$ binaries and $2{,}282$ constraints at $n=20$; $1{,}225$ binaries and $39{,}202$ constraints at $n=100$). We solve each instance with HiGHS via \texttt{scipy.optimize.milp}, capping the solver at $180$\,s with a $0.1\%$ relative-gap tolerance, and additionally wrapping every solve in a $300$\,s subprocess wall-clock cap. We generate $155$ instances per $n$ ($100$ random, $50$ correlated, $5$ reversed). Across the $542$ instances solved within the limits, the worst observed exact ratio is $1.31$, attained on a correlated configuration at $n=50$, and the LP relaxation upper-bounds the ratio by $1.32$ across $612$ instances. Together with the exhaustive enumeration up to $n=9$ (worst ratio $1.46$ over $699{,}282$ instances), this provides strong empirical evidence for the 2-approximation conjecture.

\section{Additional experimental results}\label{app:extra-experiments}

\subsection{Approximation ratio for the all-ascending case}\label{app:sorted-case-experiments}

\begin{figure}[h]
    \centering
    \includegraphics[width=.55\textwidth]{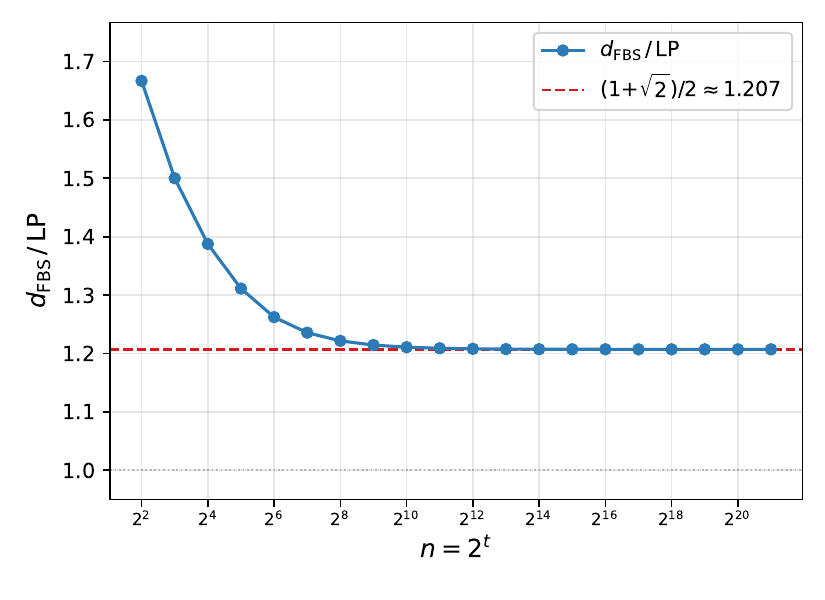}
    \caption{Ratio $d_{\mathrm{FBS}}/\mathrm{LP}$ for the all-ascending case $a=s$ with $n=2^t$, $t=2,\ldots,21$. The dashed red line marks the theoretical limit $(1{+}\sqrt{2})/2\approx 1.207$ from Proposition~\ref{prop:ascending-bound}.}
    \label{fig:sorted-case}
\end{figure}

Proposition~\ref{prop:ascending-bound} predicts that, for the all-ascending configuration $a=s$ and $n=2^t$, the ratio $d_{\mathrm{FBS}}/\mathrm{LP}$ converges to $(1{+}\sqrt{2})/2\approx 1.207$.
We compute the closed-form FBS cost (Proposition~\ref{thm:five-block}) and the LP relaxation bound for each $n=2^t$ with $t=2,\ldots,21$ (up to $n\approx 2\times 10^6$).
Figure~\ref{fig:sorted-case} shows the convergence: the ratio decreases monotonically from $1.67$ ($n{=}4$) and stabilises at $1.207$ for $t\ge 10$.

\begin{figure}[h]
    \centering
    \includegraphics[width=.55\textwidth]{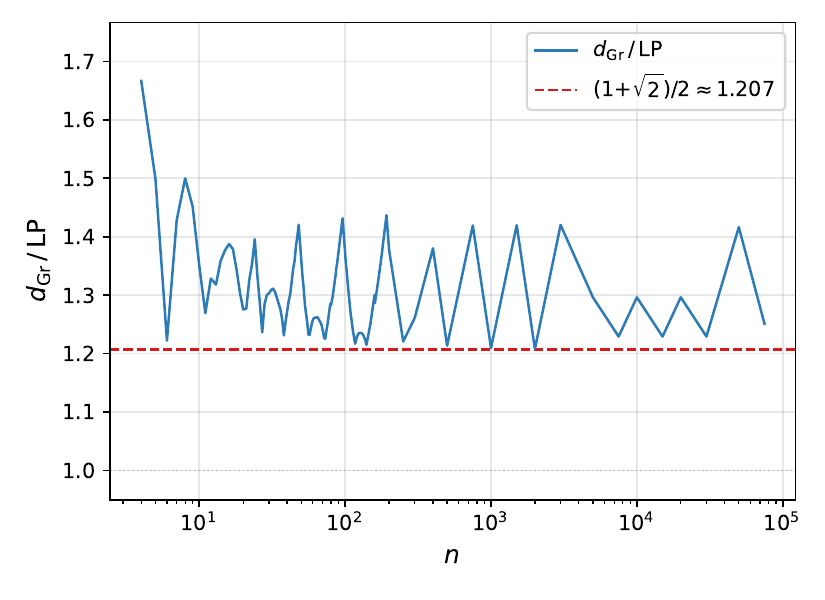}
    \caption{Ratio $d_{\mathrm{FBS}}/\mathrm{LP}$ for the all-ascending case $a=s$ with arbitrary~$n$ (dense for $n\le 200$, sparse up to $n=75{,}000$). The ratio oscillates but stays bounded; local minima near powers of two approach the $(1{+}\sqrt{2})/2$ limit (dashed red line).}
    \label{fig:sorted-case-arbitrary}
\end{figure}

For general (non-power-of-two) $n$, the five-block structure of Proposition~\ref{thm:five-block} does not apply, and the ratio exhibits a persistent oscillation.
Figure~\ref{fig:sorted-case-arbitrary} shows $d_{\mathrm{FBS}}/\mathrm{LP}$ for $n=4,\ldots,200$ (every integer) and selected values up to $n=75{,}000$, all with $a=s$ and $\Delta^{\ast}$ set to the tightest feasible value.
The local minima, occurring near powers of two, approach the $(1{+}\sqrt{2})/2$ limit, while the local maxima reach $\approx 1.42$ but remain well below~$2$.

\subsection{Synthetic baseline comparisons}\label{app:baselines-extra}

We generate synthetic instances with $n\in\{100,1000,5000\}$ and score--attribute correlation $\mathrm{corr}\in\{0, 0.5, 1\}$, averaging over $5$ repetitions.
FBS consistently traces a Pareto front that dominates all baselines: it achieves the same fairness at lower cost, or higher fairness at the same cost.
The gap is most pronounced at high correlation ($\mathrm{corr}=1$), where the discretization baselines pay a substantial cost to reach even moderate $\phi$ values, while FBS reaches $\phi=1$ along a smooth, efficient trade-off curve.
DetConstSort with more groups ($G=5,10$) reaches higher $\phi$ than with fewer groups, but at a cost that exceeds FBS for the same fairness level.
FA*IR's one-sided minimum-representation constraint provides only a modest fairness improvement, as it addresses group underrepresentation but not the symmetric balance that $\phi$ measures.
These results confirm that discretizing a continuous attribute incurs an inherent information loss: group-fair methods can only balance across bin boundaries, while FBS directly optimises the continuous fairness measure at the pairwise level.

\begin{figure}[h]
    \centering
    \includegraphics[width=\textwidth]{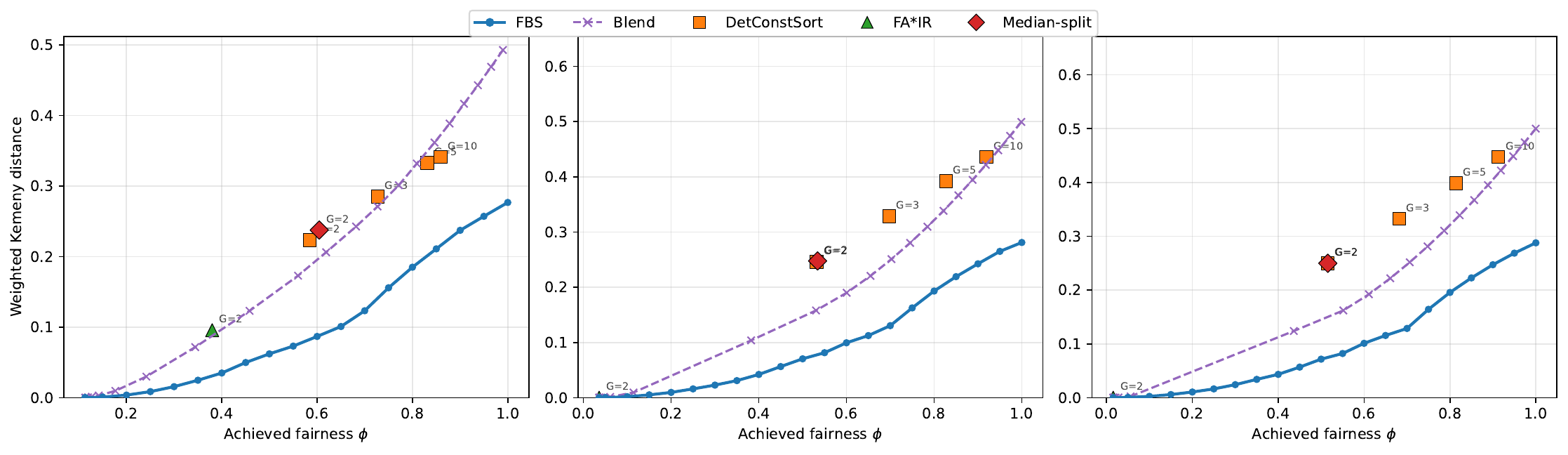}
    \caption{FBS vs.\ discretization baselines on synthetic data with $\mathrm{corr}=0.5$, $n\in\{100,1000,5000\}$. Each panel shows achieved fairness $\phi$ ($x$-axis) vs.\ weighted Kemeny distance ($y$-axis). FBS (blue line) traces a Pareto front by sweeping $\psi$; baselines are shown as individual points for each number of groups $G$. FBS dominates all baselines.}
    \label{fig:baselines}
\end{figure}

\begin{figure}[h]
    \centering
    \includegraphics[width=\textwidth]{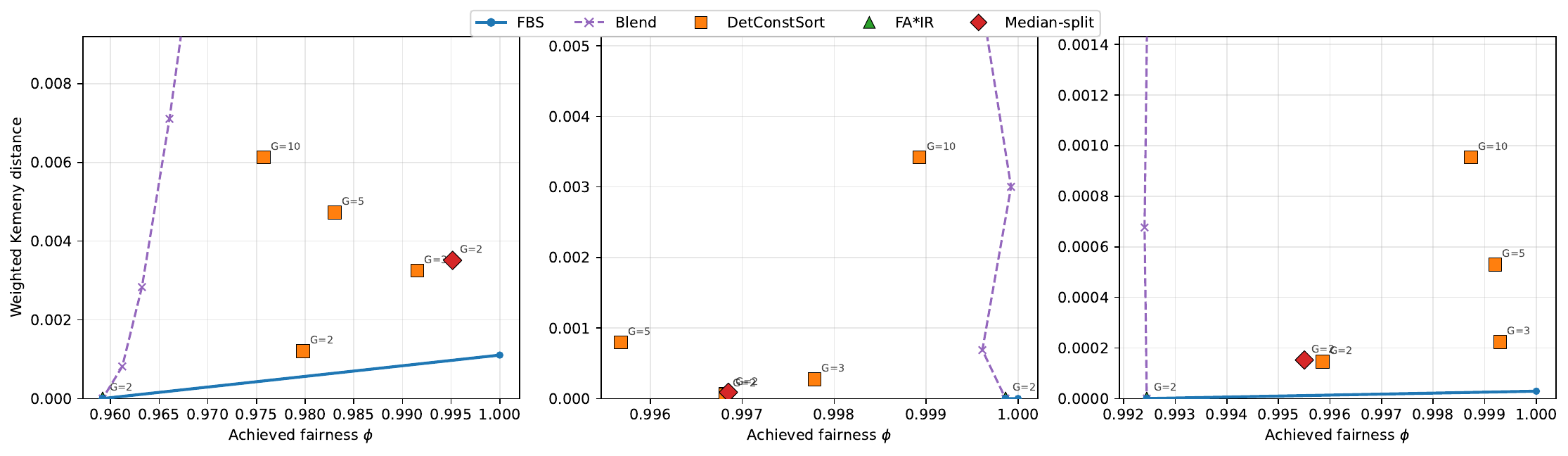}
    \caption{FBS vs.\ discretization baselines on synthetic data with $\mathrm{corr}=0$ (uncorrelated scores and attributes), $n\in\{100,1000,5000\}$.}
    \label{fig:baselines-corr0}
\end{figure}

\begin{figure}[h]
    \centering
    \includegraphics[width=\textwidth]{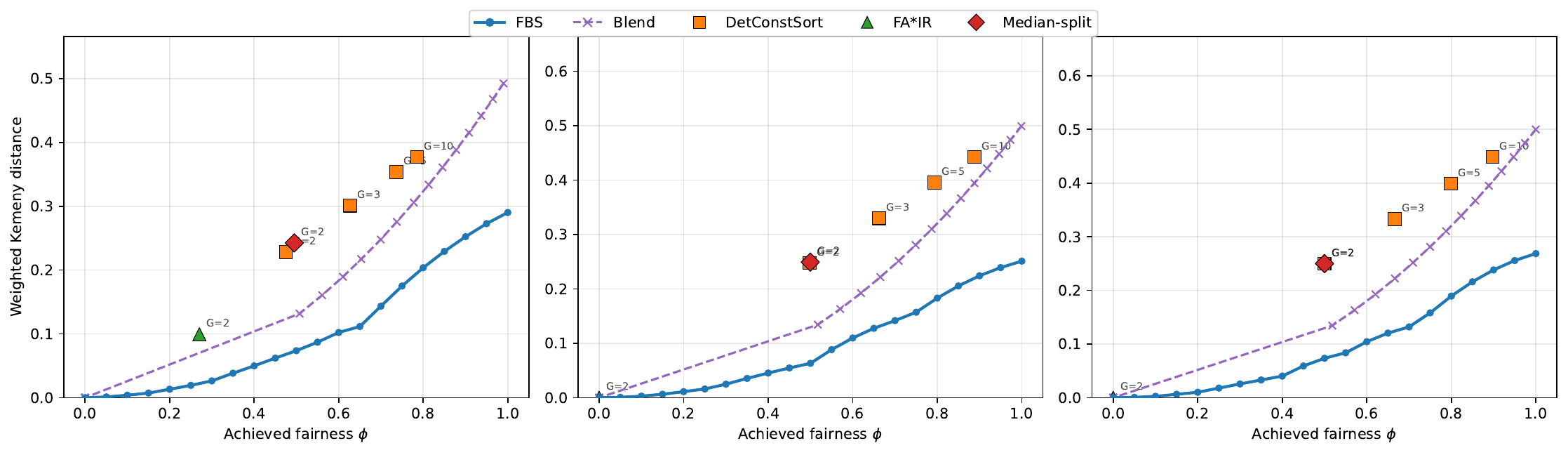}
    \caption{FBS vs.\ discretization baselines on synthetic data with $\mathrm{corr}=1$ (perfectly correlated scores and attributes), $n\in\{100,1000,5000\}$.}
    \label{fig:baselines-corr1}
\end{figure}

\clearpage
\subsection{Continuous-fairness repair baselines}\label{app:repair-baselines}

The discretization baselines of Figure~\ref{fig:baselines-real} enforce group constraints on a binned attribute. A different route to a fair ranking is to repair the score itself so that it no longer depends on the attribute, and then sort by the repaired score. We evaluate two such repair-then-sort baselines on the same data and axes as Figure~\ref{fig:baselines-real}: achieved fairness $\phi$ on the original continuous attribute, and weighted Kemeny distance from $r_s$.

\emph{Wasserstein repair.} Following the geometric repair of~\cite{feldman2015}, we bin the attribute into $G\in\{2,3,5,10\}$ equal-frequency groups and move the score distribution of each group towards a common target, the pointwise median of the groups' quantile functions: an item at quantile $q$ of its group receives the score $(1-\lambda)\,s+\lambda\,F^{-1}(q)$, where $F^{-1}$ is the target quantile function and $\lambda\in\{0.25,0.5,0.75,1\}$ is the repair strength ($\lambda=0$ leaves $r_s$ unchanged, $\lambda=1$ gives every group the same score distribution). Items are then sorted by the repaired score.

\emph{Residual repair.} In the spirit of~\cite{johndrow2019algorithm}, this variant works on the continuous attribute directly. We estimate the conditional mean of the score given the attribute, $\hat f(a)$, by a centred moving average of the scores over the items sorted by attribute (window of $\max(11,\lfloor n/50\rfloor)$ items), and compute the residual $s-\hat f(a)$. Items are sorted by $(1-\lambda)\,\tilde s+\lambda\,\tilde e$, where $\tilde s$ and $\tilde e$ are the standardized score and residual and $\lambda$ takes the same values as above; $\lambda=1$ ranks by the residual alone, i.e.\ removes the local influence of the attribute on the score.

\begin{figure}[h]
    \centering
    \includegraphics[width=\textwidth]{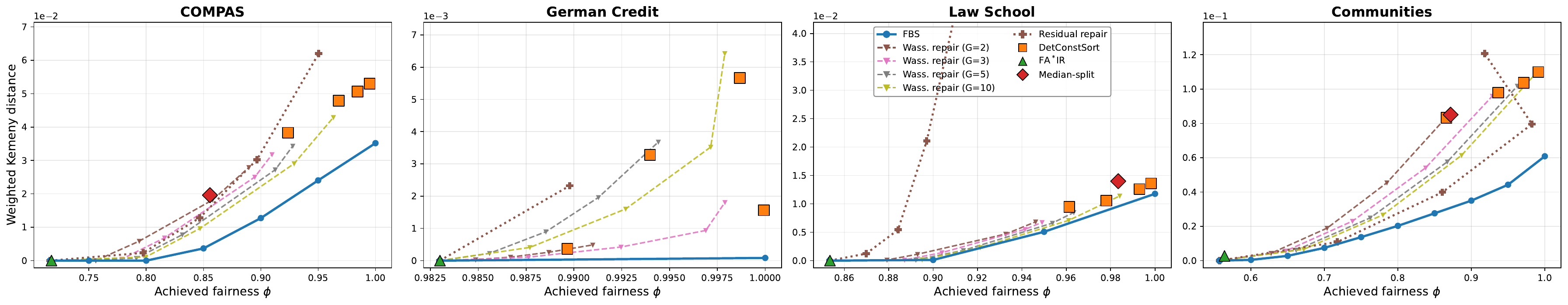}
    \caption{FBS vs.\ repair-then-sort baselines on the real datasets; axes as in Figure~\ref{fig:baselines-real}. Dashed lines: Wasserstein repair for $G\in\{2,3,5,10\}$, sweeping the strength $\lambda$; dotted line with crosses: residual repair, sweeping $\lambda$; the discretization baselines of Figure~\ref{fig:baselines-real} are shown as points for reference. Each panel shows the rankings at least as fair as $r_s$; on German Credit, three of the four residual-repair rankings are less fair than $r_s$, since the repair removes the influence of the attribute on the score rather than targeting $\phi$, and are omitted. On Law School the $y$-axis is cut at three times the largest cost of the other methods; the residual repair continues beyond it, up to a cost of $0.20$ at $\phi=1$ (Table~\ref{tab:repair}).}
    \label{fig:baselines-repair}
\end{figure}

Figure~\ref{fig:baselines-repair} overlays the two families on the curves of Figure~\ref{fig:baselines-real}. For each repair ranking whose fairness lies within the range spanned by FBS, we compute the ratio between its cost and the cost of FBS at the same fairness, obtained by linear interpolation of the FBS curve; a ratio above one means that FBS reaches the same fairness at lower cost. Table~\ref{tab:repair} reports, for each family and dataset, the fairest ranking it produces and the median and maximum ratio. FBS is cheaper at every matched point on all four datasets. The Wasserstein repair comes closest (smallest ratio $1.05$, on Law School; median ratios between $1.3$ and $2.1$ outside German Credit); the residual repair pays $1.4$ to $2.6$ times the FBS cost on COMPAS and Communities and far more on Law School, where removing the influence of the attribute reorders a large part of the list. The large ratios on German Credit reflect its near-fair starting point ($\phi_0=0.98$), where FBS reaches $\phi=1$ at a cost of about $10^{-4}$.

\begin{table}[h]
\centering
{\setlength{\tabcolsep}{3pt}
\begin{tabular}{@{}llcccc@{}}
\toprule
Baseline & Dataset & $\phi_0$ & $\phi$ reached & cost vs.\ FBS (median) & cost vs.\ FBS (max) \\
\midrule
Wasserstein repair & COMPAS & 0.72 & 0.96 & 2.1$\times$ & 5.8$\times$ \\
 & German Credit & 0.98 & 1.00 & 18$\times$ & 88$\times$ \\
 & Law School & 0.85 & 0.98 & 1.3$\times$ & 11$\times$ \\
 & Communities & 0.56 & 0.99 & 2.0$\times$ & 2.8$\times$ \\
\midrule
Residual repair & COMPAS & 0.72 & 0.95 & 2.6$\times$ & 3.8$\times$ \\
 & German Credit & 0.98 & 0.99 & 70$\times$ & 70$\times$ \\
 & Law School & 0.85 & 1.00 & 49$\times$ & 189$\times$ \\
 & Communities & 0.56 & 0.98 & 1.4$\times$ & 3.2$\times$ \\
\bottomrule
\end{tabular}}
\caption{Repair-then-sort baselines vs.\ FBS at matched fairness. ``$\phi$ reached'' is the fairest ranking the family produces; the cost ratio is the repair cost divided by the FBS cost at the same $\phi$, with median and maximum taken over the family's rankings within the fairness range of FBS.}
\label{tab:repair}
\end{table}

\subsection{Ranking quality beyond the Kemeny distance: NDCG and FBS-NDCG}\label{app:ndcg}

The weighted Kemeny distance charges every inverted pair by its score gap, wherever the pair sits in the list. NDCG with the original score as relevance is a second utility measure that weighs the top of the list more heavily: $r_s$ is the ideal ranking and has NDCG equal to one, the relevance of an item is its score (linear gain), and position $p$ is discounted by $\delta_p=1/\log_2(p+1)$. We report NDCG over the full list and truncated at the top $10$ and top $100$ positions.

\emph{FBS as a cost-agnostic framework.} At each step, FBS performs the cheapest fairness-improving adjacent swap under a per-swap cost, and any utility whose change under one adjacent swap is cheap to evaluate can serve as this cost. Algorithm~\ref{alg:fbs-weighted} uses the score gap $|S[k]-S[k+1]|$ of the two items as the cost and minimizes the weighted Kemeny distance, with the guarantees of Theorem~\ref{thm:unweighted-optimality} and Proposition~\ref{prop:weighted-approx}. Replacing the key of the heap by the decrease of DCG caused by the swap, $(S[k]-S[k+1])(\delta_k-\delta_{k+1})$, makes swaps near the top of the list expensive, so the greedy protects the head of the list; we call this variant FBS-NDCG. Both variants perform exactly $K^{\ast}$ majority-sign swaps and reach the same fairness; they differ only in which swaps they choose, that is, in where the displacement is spent. FBS-NDCG is a greedy heuristic for NDCG.

Table~\ref{tab:ndcg} reports NDCG at matched fairness. For each dataset we take the discretization baseline of Figure~\ref{fig:baselines-real} with the highest achieved fairness (DetConstSort with $G=10$; $G=3$ on German Credit) and evaluate FBS and FBS-NDCG at the same $\phi$ by linear interpolation along their $\psi$-sweeps ($\psi$ from $0$ to $1$ in steps of $0.1$). As a reference we also report a uniformly random permutation, averaged over $10$ draws: it has $\phi\approx 1$ but is otherwise arbitrary, so its NDCG calibrates the scale. Two observations follow. First, fair rankings remain practical: the random fair ranking loses most of its top-heavy utility (NDCG@10 between $0.18$ and $0.78$), while FBS keeps NDCG@10 between $0.82$ and $1.00$ and full-list NDCG at or above $0.98$ on every dataset. Second, FBS keeps the full-list displacement small but gives up some top-$k$ utility, most visibly on the datasets with coarse scores (COMPAS, Law School), where it falls below the best baseline on NDCG@10 and NDCG@100. This is exactly the regime where FBS-NDCG, by pricing swaps near the top higher, recovers NDCG@10 and NDCG@100 of $1.00$ at the same fairness, matching or exceeding the best baseline on every dataset and metric. Targeting top-$k$ utility therefore requires only the NDCG cost inside the same routine.

\begin{table}[h]
\centering
{
\begin{tabular}{@{}llcccc@{}}
\toprule
Dataset (matched $\phi$) & Metric & Random fair & Best baseline & FBS & FBS-NDCG \\
\midrule
COMPAS ($0.995$) & NDCG & 0.920 & 0.995 & 0.991 & 0.998 \\
 & NDCG@10 & 0.400 & 1.000 & 0.889 & 1.000 \\
 & NDCG@100 & 0.440 & 0.997 & 0.872 & 1.000 \\
\midrule
German Credit ($1.000$) & NDCG & 0.790 & 0.999 & 1.000 & 1.000 \\
 & NDCG@10 & 0.210 & 0.992 & 1.000 & 1.000 \\
 & NDCG@100 & 0.290 & 0.996 & 1.000 & 1.000 \\
\midrule
Law School ($0.998$) & NDCG & 0.980 & 1.000 & 0.999 & 1.000 \\
 & NDCG@10 & 0.780 & 1.000 & 0.932 & 1.000 \\
 & NDCG@100 & 0.780 & 1.000 & 0.951 & 1.000 \\
\midrule
Communities ($0.991$) & NDCG & 0.810 & 0.951 & 0.980 & 0.995 \\
 & NDCG@10 & 0.180 & 0.944 & 0.826 & 1.000 \\
 & NDCG@100 & 0.240 & 0.753 & 0.943 & 1.000 \\
\bottomrule
\end{tabular}}
\caption{NDCG at matched fairness, with the original score as relevance (NDCG $=1$ for $r_s$). For each dataset, the matched $\phi$ is the highest fairness reached by a discretization baseline (DetConstSort with $G=10$, or $G=3$ on German Credit); ``Best baseline'' is that baseline, FBS and FBS-NDCG are evaluated at the same $\phi$, and ``Random fair'' is a uniformly random permutation ($\phi\approx 1$), averaged over $10$ draws.}
\label{tab:ndcg}
\end{table}

\subsection{Other measures of dependence between position and attribute}\label{app:dependence}

CPP measures monotone dependence between position and attribute (Appendix~\ref{app:motivations}). To check whether FBS also reduces dependence as measured by other statistics, we evaluate four measures between the position of an item in the ranking and its attribute value, all of which are zero under independence: the absolute Spearman correlation; distance correlation~\cite{szekely2007measuring}; the Hilbert--Schmidt independence criterion (HSIC)~\cite{gretton2005measuring} with Gaussian kernels whose bandwidth is the median pairwise distance, normalised by the geometric mean of the two self-HSIC terms so that it lies in $[0,1]$; and a positional discrepancy, the largest one-dimensional Wasserstein distance between the attribute distributions of the top, middle and bottom thirds of the ranking, divided by the standard deviation of the attribute. The last three also detect non-monotone dependence. Distance correlation and HSIC are quadratic in $n$, so on Law School ($n=18{,}692$) they are computed on one fixed random subsample of $8{,}000$ items, the same for every ranking of that dataset; the other two measures use all items.

Table~\ref{tab:dependence} reports the four measures for the score ranking $r_s$ and for the FBS ranking at full fairness ($\psi=1$). On the three datasets whose initial ranking is far from fair (Communities, COMPAS, Law School), FBS reduces all four measures substantially, by factors between $1.6$ and $11$; on German Credit the four measures are already near zero at $r_s$ and change only negligibly. FBS therefore improves fairness as measured by these statistics as well. Some non-monotone dependence remains at $\psi=1$ (distance correlation and HSIC stay above zero), consistent with CPP measuring monotone dependence.

\begin{table}[h]
\centering
{\setlength{\tabcolsep}{3pt}
\begin{tabular}{@{}lccccc@{}}
\toprule
Dataset & $\phi_0$ & $|$Spearman $\rho|$ & distance correlation & HSIC & positional discrepancy \\
\midrule
Communities & 0.56 & 0.62 $\to$ 0.19 & 0.57 $\to$ 0.36 & 0.300 $\to$ 0.143 & 1.26 $\to$ 0.40 \\
COMPAS & 0.72 & 0.42 $\to$ 0.13 & 0.41 $\to$ 0.23 & 0.131 $\to$ 0.042 & 0.94 $\to$ 0.48 \\
Law School & 0.85 & 0.22 $\to$ 0.06 & 0.20 $\to$ 0.10 & 0.032 $\to$ 0.003 & 0.47 $\to$ 0.15 \\
German Credit & 0.98 & 0.03 $\to$ 0.01 & 0.06 $\to$ 0.07 & 0.005 $\to$ 0.005 & 0.16 $\to$ 0.17 \\
\bottomrule
\end{tabular}}
\caption{Dependence between rank position and attribute for the score ranking $r_s$ (left of each arrow) and for the FBS ranking at $\psi=1$ (right); lower means less dependence. Distance correlation and HSIC on Law School are computed on a fixed subsample of $8{,}000$ items.}
\label{tab:dependence}
\end{table}

\clearpage
\section{Hardware,  Software and Code}\label{app:hardware}

All experiments were run on a single machine with the following specifications:

\begin{itemize}
\item \textbf{Hardware:} Apple MacBook Pro with Apple M1 Pro chip, 8-core CPU, 16\,GB unified memory.
\item \textbf{OS:} macOS 15.6.1 (Sequoia).
\item \textbf{Code:} The source code to run FBS and to reproduce the experiments is publicly available at \url{https://github.com/Ambress92/FairBubbleSort}.
\end{itemize}


\end{document}